%% file: main.tex
\documentclass[10pt,twocolumn,letterpaper]{article}

\usepackage[pagenumbers]{cvpr} 

\usepackage{graphicx}
\usepackage{amsmath}
\usepackage{amssymb}
\usepackage{booktabs}
\usepackage{changes}
\usepackage[percent]{overpic}
\setdeletedmarkup{\textcolor{red}{\sout{#1}}}

\definecolor{cvprblue}{rgb}{0.21,0.49,0.74}
\usepackage[pagebackref,breaklinks,colorlinks,citecolor=cvprblue]{hyperref}

\usepackage[capitalize]{cleveref}
\usepackage{orcidlink}
\usepackage{bm}
\usepackage{cancel}
\crefname{section}{Sec.}{Secs.}
\Crefname{section}{Section}{Sections}
\Crefname{table}{Table}{Tables}
\crefname{table}{Tab.}{Tabs.}

\def\paperID{389}
\def\confName{3DV\xspace}
\def\confYear{2026\xspace}

\newcommand{\algname}{SC-Diff}

\title{\algname: 3D Shape Completion with Latent Diffusion Models}

\author{
Simon Schaefer*$^{1,2,3}$\\
\small simon.k.schaefer@tum.de
\and
Juan D. Galvis*$^{1}$\\
\small juan.galvis@tum.de
\and
Xingxing Zuo$^{4,\dagger}$\\
\small xingxing.zuo@mbzuai.ac.ae
\and
Stefan Leutenegger$^{1,2,3}$\\
\small lestefan@ethz.ch
\thanks{$^{1}$Technical University of Munich}
\thanks{$^{2}$Munich Center for Machine Learning (MCML)}
\thanks{$^{3}$ETH Zurich}
\thanks{$^{4}$MBZUAI}
\thanks{$*$S. Schaefer and J. Galvis are co-first authors.}
\thanks{$^{\dagger}$X. Zuo is the corresponding author.}}
\renewcommand\footnotemark{}  

\begin{document}
\maketitle

\begin{abstract}
We present a novel 3D shape completion framework that unifies multimodal conditioning, leveraging both 2D images and 3D partial scans through a latent diffusion model. Shapes are represented as Truncated Signed Distance Functions (TSDFs) and encoded into a discrete latent space jointly supervised by 2D and 3D cues, enabling efficient high-resolution processing while reducing GPU memory usage by 30\% compared to state-of-the-art methods. Our approach guides the generation process with flexible multimodal conditioning, ensuring consistent integration of 2D and 3D information from encoding to reconstruction. Our training strategy simulates realistic partial observations, avoiding assumptions about input structure and improving robustness in real-world scenarios. Leveraging our efficient latent space and multimodal conditioning, our model generalizes across object categories, outperforming class-specific models by 12\% and class-agnostic models by 47\% in $l_1$ reconstruction error, while producing more diverse, realistic, and high-fidelity completions than prior approaches.
\end{abstract}

\vspace{-0.3cm}
\input{chapters/01_introduction}
\input{chapters/02_related_work}
\input{chapters/03_method}
\input{chapters/04_experiments}

\input{chapters/05_conclusion}

\newpage
{
    \small
    \bibliographystyle{ieeenat_fullname}
    \bibliography{main}
}

\input{chapters/X_supplementary}

\end{document}

%% file: chapters/01_introduction.tex
\section{Introduction}
\label{section:introduction}
The rapid evolution of digital technologies has made 3D content a cornerstone in augmented/virtual reality, gaming, robotics, and industrial design. These applications rely on complete and accurate 3D models, but creating such models remains challenging and labor-intensive. Commodity RGB-D sensors (e.g., Intel RealSense, Microsoft Kinect, iPhone LiDAR) have enabled significant progress in capturing real-world 3D data. However, despite advancements in reconstruction quality \cite{azinovic2022neuralrgbdreconstruction,park2019deepsdf}, the acquired scans are typically partial, incomplete, or noisy due to occlusions and limited sensor viewpoints, leaving large gaps in the reconstruction pipeline. Addressing this fundamental challenge is essential for many downstream technologies.

\begin{figure}[t]
  \centering
  \includegraphics[width=\linewidth]{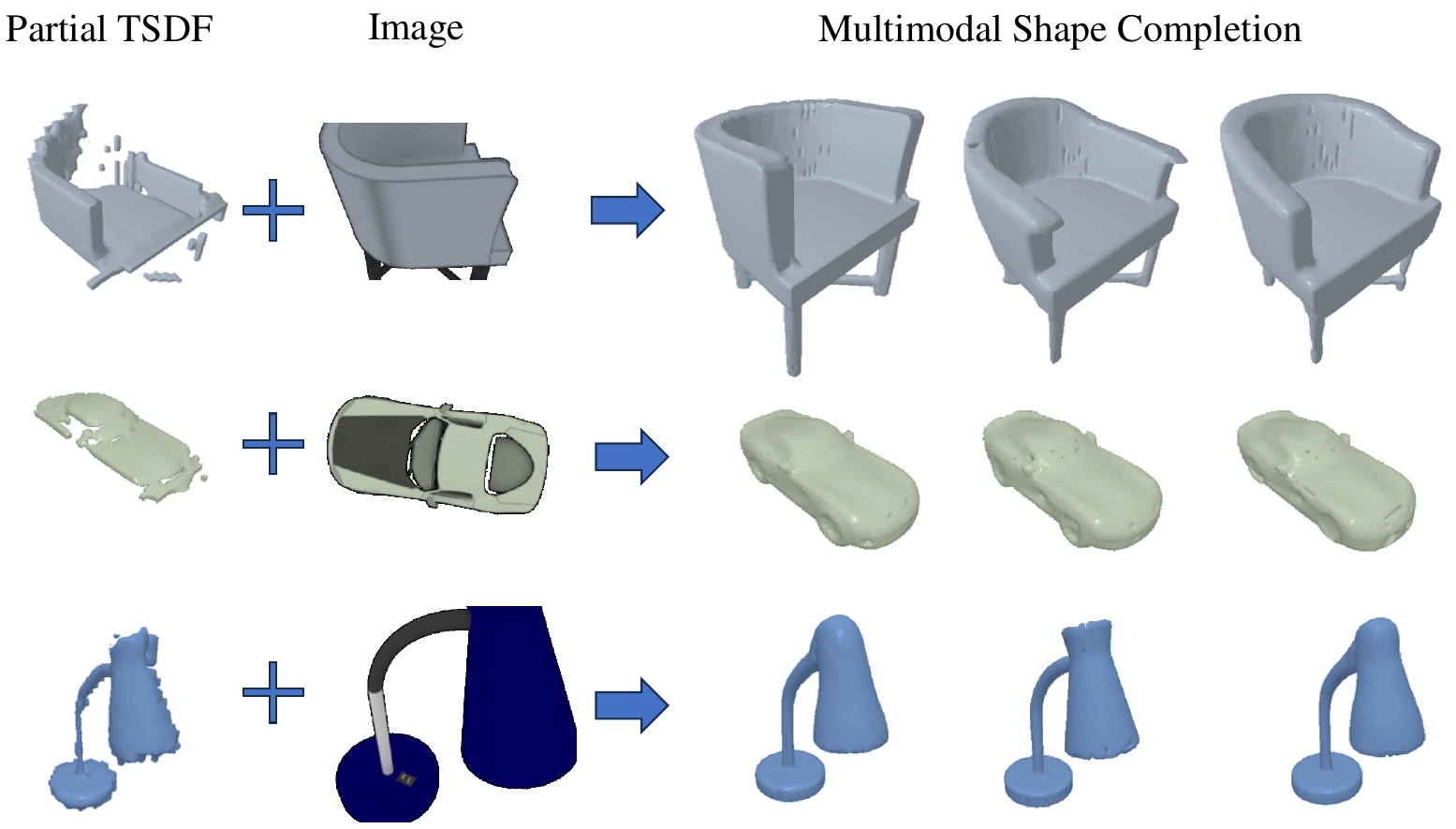}
  \caption{Our method, \algname, is a latent diffusion framework for 3D shape completion that leverages 2D and 3D conditioning across the entire pipeline, scales to high-resolution TSDFs, and generalizes across object categories, achieving state-of-the-art accuracy, diversity, and realism.}
  \label{fig:results_3depn1}
\end{figure}

A central design choice in tackling 3D shape completion is the representation of 3D geometry. Signed Distance Functions (SDFs), meshes, point clouds, and neural radiance fields each offer unique advantages, but SDF-based volumetric representations remain a robust option for high-fidelity completion. Traditional learning-based completion methods~\cite{dai2017shape_3depn,yan2022shapeformer} often map partial scans directly to full shapes. While effective to some extent, such deterministic mappings tend to overfit and fail to capture the inherent ambiguity of missing geometry. Generative approaches instead model shape completion as a conditional generation problem, yielding more diverse and plausible results~\cite{mittal2022autosdf,zhang2021unsupervised_gan_completion,cheng2023sdfusion,chu2023diffcomplete}. Diffusion models in particular have emerged as powerful probabilistic generative frameworks, offering high-quality and diverse outputs~\cite{ho2020ddpm,rombach2022latentdiffusionmodels}.

Despite this progress, existing diffusion-based shape completion methods face several critical limitations. 
First, they typically adopt simplified training setups such as filling artificially cropped cubic regions~\cite{cheng2023sdfusion}, which do not reflect realistic sensor-induced incompleteness. 
Second, they do not scale to high resolutions. Current state-of-the-art methods operate at $32^3$ voxel grids~\cite{cheng2023sdfusion,chu2023diffcomplete} since dense latent grids and memory-hungry architectures prohibit scaling to larger volumes. As a result, fine-grained details are lost, and applying these methods to real-world, high-resolution 3D scans remains infeasible.
Third, they often operate under restrictive unimodal conditioning~\cite{chu2023diffcomplete,mittal2022autosdf,rao2022patchcomplete} or assume multi-view inputs, which largely limits real-world applications, as obtaining sufficient coverage to capture geometric detail is infeasible and comes with extensive computational and memory costs~\cite{gao2024cat3d}.
To overcome these challenges, we propose \algname, a framework for realistic, multimodal, and scalable 3D shape completion. Our goal is to generate high-fidelity, complete shapes while integrating multimodal inputs throughout the pipeline. Specifically, we complete Truncated Signed Distance Function (TSDF) representations using a latent diffusion model conditioned on both image and spatial information.
Our contributions are:

\begin{itemize}
\item We are the first to unify latent diffusion with both 2D and 3D conditioning, applied consistently from encoder supervision to the denoising process, for object completion. We introduce a discrete latent representation jointly trained with 2D and 3D supervision, compressing TSDF volumes into a compact space. This enables scaling up to $64^3$ resolution completion while reducing GPU memory usage by $30\%$ compared to state-of-the-art~\cite{chu2023diffcomplete, cheng2023sdfusion}.

\item We introduce a guidance mechanism that combines cross-attention with ControlNet-inspired conditioning, enabling flexible integration of complementary modalities. We introduce a training strategy that simulates realistic partial shapes directly from 3D datasets, better reflecting real-world occlusions and noise while not making any assumptions about the input's structure. This multimodal approach enables class-agnostic generation while producing more realistic, diverse, and high-fidelity completions compared to prior methods.

\item We achieve state-of-the-art performance on both synthetic and real-world data, outperforming class-specific models by $12\%$ and class-agnostic models by $47\%$ in $l_1$ reconstruction error.
\end{itemize}

%% file: chapters/02_related_work.tex
\section{Related Work}
\label{section:related_work}
Traditional approaches to 3D shape completion primarily focus on repairing minor imperfections and geometrical gaps, employing strategies like Laplacian hole filling \cite{li2012temporally_laplacian} and Poisson surface reconstruction \cite{centin2015mesh_poisson}. While effective for filling small gaps in 3D shapes, these methods are of limited use in scenarios with larger gaps in the input scans. However, the advent of extensive 3D datasets has led to the development of learning-based approaches. While retrieval-based methods \cite{bosche2008automated_retrieval,zhang2019view_retrieval,beyer2022weakly_cad_retrieval} extract the most fitting shapes from a database to complete the incomplete input, learning-based fitting methods \cite{yuan2018pc_pcn,dai2017_3depn,dai2020sg_nn,dai2021spsg, binbin2022completion} that aim to minimize the disparity between network-predicted shapes and actual ground truth have become more prevalent. For example, 3D-EPN~\cite{dai2017_3depn} and Scan2Mesh~\cite{dai2019scan2mesh} have employed 3D encoder-decoder architectures for predicting complete shapes from partial volumetric data. More recently, PatchComplete~\cite{rao2022patchcomplete} introduces a novel design by learning multi-resolution patch priors, enhancing the completion of shapes, particularly for previously unseen categories. 
While these models can provide high-accuracy completions, their deterministic nature may lead to overfitting. Furthermore, they often lack the ability to generate finer details in shapes.

Generative methods, including Generative Adversarial Networks (GANs) \cite{zhang2021unsupervised_gan_completion} and AutoEncoders \cite{mittal2022autosdf}, have recently been introduced to the field. These methods excel in generating a diverse range of plausible shapes from partial inputs. However, this often comes at the cost of accuracy, especially when a precise ground truth is available for training. 
Diffusion models have emerged as a popular family of generative models, renowned for their impressive quality, diversity, and expressiveness in tasks such as image synthesis \cite{rombach2022latentdiffusionmodels}, image editing \cite{meng2021sdedit}, and text-to-image synthesis \cite{ramesh2022hierarchical_dalle2}. While extensively researched for 2D data, their application to 3D data has only been explored very recently, using different 3D representations such as point clouds \cite{lyu2021pc_diff_conditional,zhou20213dpc_diff_pointvoxel}, neural radiance fields \cite{metzer2022latentnerf}, or signed distance functions \cite{chou2023diffusionsdf,li2023generalizedsdfdiff}. 
SDFusion~\cite{cheng2023sdfusion}, specifically, introduces a latent diffusion model for shape generation aimed at mitigating the computational challenges associated with 3D representations, and conditions it on various modalities like images, text, and shapes. However, SDFusion frames shape completion as filling simple, artificially removed cubic regions, rather than reconstructing the irregular, realistic missing parts encountered in real-world scans. In contrast, our approach does not make any assumption on the partial shape as real-world scans are characterized by noise and varying degrees of incompleteness. 
Another approach, DiffComplete~\cite{chu2023diffcomplete}, offers a more realistic shape completion method but directly works on the 3D space and is therefore restricted to lower-resolution voxel grids of size $32^3$. As opposed to DiffComplete, our approach can achieve much higher resolution by encoding the 3D space to compact latent codes.

More recently, object completion has also been explored in a zero-shot setting using foundation models by using score distillation sampling~\cite{kasten2023point}. However, these approaches require a lengthy per-object optimization process during test-time and rely on sparse yet mostly complete projections from a single viewpoint \cite{li2025genpc, kasten2023point}. Similarly, Gao et al.~\cite{gao2024cat3d} propose multi-view completion using a diffusion model conditioned on multiple images. While promising, both directions make strong assumptions about the input data and incur substantial computational and memory costs. In contrast, our model makes no assumptions on the structure of its conditioning signal and remains comparatively lightweight in memory usage.
Furthermore, most previous methods \cite{rao2022patchcomplete,muller2023diffrf,metzer2022latentnerf} work on the assumption that the object class is known prior to the completion, which makes them hard to deploy in real-world applications. In contrast, our model is class-agnostic, and outputs high-detailed, realistic completions, even for unseen classes. 

%% file: chapters/03_method.tex
\begin{figure*}[!htb]
\centering
\includegraphics[width=360pt]{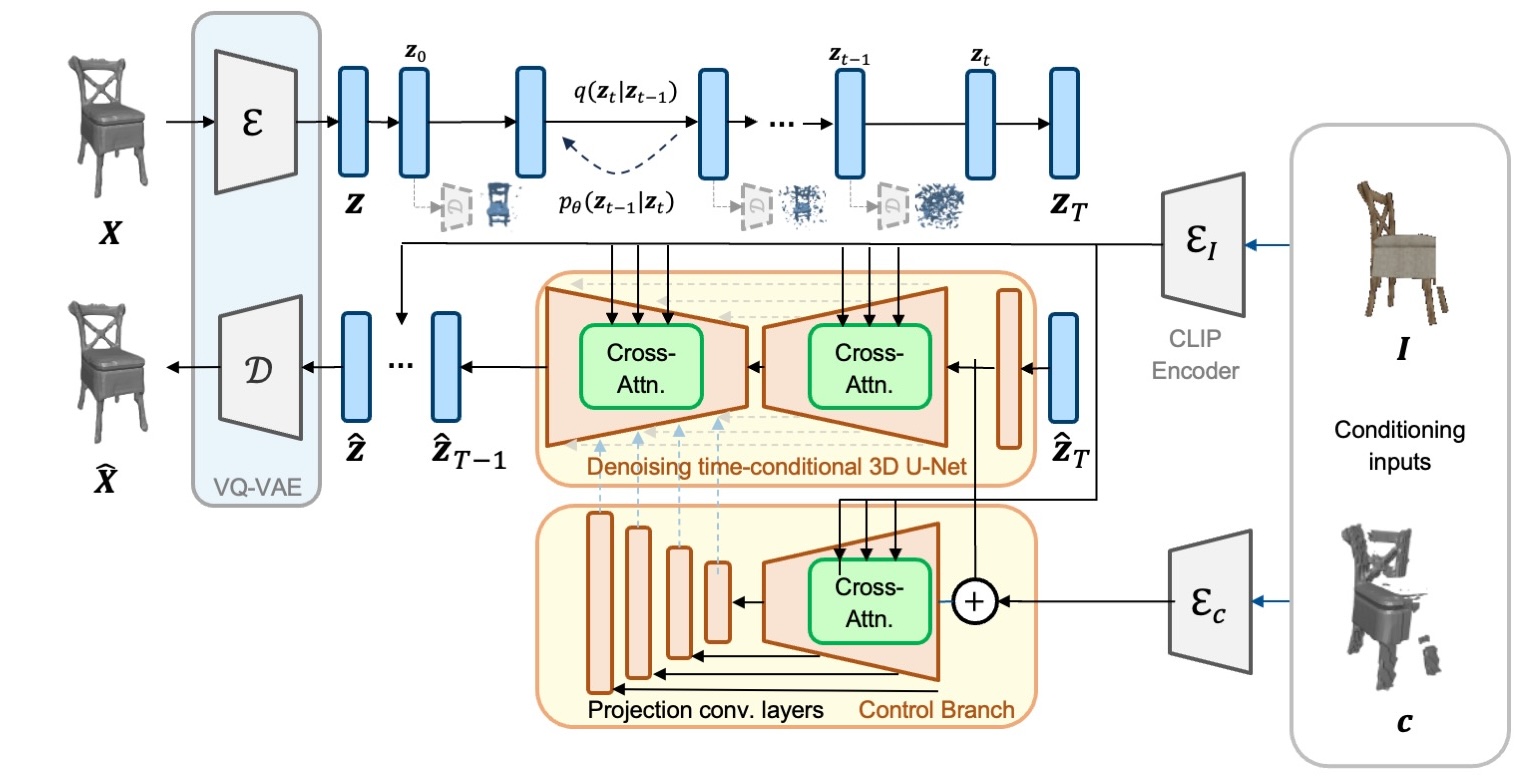}
\vspace{-1em}
\caption[Overview of the Latent Shape Completion Diffusion Model]{We train a Vector Quantized Variational Autoencoder (VQ-VAE) to encode complete TSDF voxel grids $\mathbf{X}$ into a compact latent variable $\mathbf{z}$. Then, we use this learned latent space to train our conditional shape completion diffusion model. We corrupt the clean latent code $\mathbf{z}$ through a forward diffusion process $q (\mathbf{z}_{t}|\mathbf{z}_{-1})$ by gradually adding Gaussian noise until reaching the normally distributed variable $\mathbf{z}_T$. Our diffusion model is then trained to learn the reverse diffusion process $p_{\theta} (\mathbf{z}_{t-1}|\mathbf{z}_t)$ conditioned on a partial TSDF scan $\textbf{c}$ of the shape by aggregating features from a control branch and/or a partial RGB image $\mathbf{I}$ of the shape by applying cross-attention on CLIP features of the given images. This process allows us to recover the latent code $\mathbf{\hat{z}}$, and finally, using the VQ-VAE's decoder, the completed shape $\mathbf{\hat{X}}$. 
}\label{fig:overview}
\vspace{-1em}
\end{figure*}

\section{Methodology}
\label{section:methodology}
We approach 3D shape completion as a generative task based on a diffusion probabilistic model, aiming to produce a complete shape $\hat{\mathbf{X}}$ from a given partial scan represented by an RGB Image $\mathbf{I}$ and a partial 3D TSDF scan $\mathbf{c}$. While diffusion models have proven highly effective in generating high-resolution images, their direct application to high-resolution 3D shapes is hindered by the substantial demands on computation and memory. Consequently, 3D diffusion-based approaches are often confined to low-resolution voxel grids, typically around $32^3$ ~\cite{cheng2023sdfusion,mittal2022autosdf}. To address this limitation, we first compress the 3D shape into a compact latent space using a Vector Quantized Variational Autoencoder (VQ-VAE), as detailed in \cref{section:method_compression}. This VQ-VAE is trained with supervision from 3D TSDF as well as 2D depth and normal renderings from a raycasting-based volume rendering approach \cite{dai2021spsg}. This compression strategy allows us to implement our diffusion model in a lower-dimensional space, a process referred to as latent diffusion (\cref{section:method_latent_diff}). Our model includes two distinct conditioning mechanisms (\cref{section:method_conditioning}) in the generative process: the first leverages RGB Image $\mathbf{I}$'s CLIP features~\cite{radford2021_clip} via cross attention, and the second uses spatially localized conditions~\cite{zhang2023controlnet,chu2023diffcomplete} derived from the partial 3D TSDF scan $\mathbf{c}$. An overview of our approach is provided in \cref{fig:overview}.

\subsection{3D Shape Compression}
\label{section:method_compression}
Processing high-resolution 3D shapes poses significant computational demands. To accommodate this in the diffusion model, we compress the TSDF-based 3D shape representation $\mathbf{X} \in \mathbb{R}^{S \times S \times S}$ into a lower-dimensional latent space $\mathbf{z} \in \mathbb{R}^{D \times S_l \times S_l \times S_l}$, where $D$ and $S_l$ represent the feature size and the downsampled spatial size. As shown by the seminal work on 2D image generation using latent diffusion models \cite{rombach2022latentdiffusionmodels}, the compression preprocessing before generation allows our diffusion model to focus more effectively on the generative aspect of the shape completion process. 

We employ a 3D variant of the VQ-VAE~\cite{van2017vqvae1,razavi2019vqvae} to condense the latent representations of 3D shapes. The architecture of the VQ-VAE consists of an encoder $\mathcal{E}$, a decoder $\mathcal{D}$, and a codebook $\bm{\mathcal{Z}} \in \mathbb{R}^{K_{\mathcal{Z}}\times D}$, where $K_{\mathcal{Z}}$ denotes the codebook size and $D$ the dimensionality of each latent vector.
Compared to continuous latent spaces, discrete representations offer two major advantages in our setting. First, following prior work on latent diffusion models~\cite{rombach2022ldm}, vector quantization prevents the diffusion process from drifting into regions corresponding to invalid or implausible 3D shapes. Unlike VAEs, which tend to over-regularize and over-smooth latent codes, VQ-VAEs preserves the diversity of distinctly different shape categories, allowing the model to represent sharp structural variations more faithfully. Second, VQ-VAE substantially reduces memory usage during training, since only discrete codebook indices need to be stored instead of full-precision latent feature volumes. This two-stage compression makes high-resolution volumetric completion feasible, enabling us to scale to $64^3$ voxel grids with significantly lower memory requirements than continuous alternatives.

For a given TSDF volume $\mathbf{X}$, encoding into a latent representation is done as $\mathbf{z} = \mathcal{E}(\mathbf{X})$. The decoder then reconstructs $\mathbf{X'} = \mathcal{D}(Q(\mathbf{z}))$ from $\mathbf{z}$, where $Q(\cdot)$ represents the quantization step, mapping each of the latent vectors $\mathbf{z}_i$ in $\mathbf{z}$ to the nearest codebook vector $\bm{\mathcal{Z}}_j$ using $Q(\mathbf{z}_i) = \bm{\mathcal{Z}}_j$, where $j = \arg \min_{k \in [1, K_z]} ||\mathbf{z}_i - \bm{\mathcal{Z}}_k||_2$. Here, $\bm{\mathcal{Z}}_k$ denotes the $k$-th element of $\bm{\mathcal{Z}}$. We pre-train our VQ-VAE loss with the following 3D losses commonly used in the field: the reconstruction loss $\mathcal{L}_R$, the codebook loss $\mathcal{L}_{CB}$, and commitment loss $\mathcal{L}_C$, following \cite{van2017vqvae1}.

Alongside 3D losses, our methodology incorporates 2D supervision from rendering the reconstructed TSDF grids $\mathbf{X'}$, similar to SPSG~\cite{dai2021spsg}. We render both depth ($\mathbf{D}_v'$) and normal images ($\mathbf{N}_v'$) from $\mathbf{X'}$ at each camera view $v$ in a differentiable manner via raycasting-based volume rendering.
The 2D supervision losses employed for VQ-VAE training include the reconstruction loss and adversarial loss.
The 2D reconstruction loss is an $l_1$ loss $\mathcal{L}_{R_{2D}}$ guiding depth and normal reconstruction:
\begin{equation}
\mathcal{L}_{R_{2D}} = ||\mathbf{D}_v - \mathbf{D}_v'||_1 + ||\mathbf{N}_v - \mathbf{N}_v'||_1,
\end{equation}
where $\mathbf{D}_v$ and $\mathbf{N}_v$ correspond to the ground-truth depth and normal values, and $\mathbf{D}_v'$ and $\mathbf{N}_v'$ are the rendered depth and normal images.
In addition, we apply a patch-based adversarial loss $L_{A_{2D}}$ on normals $\mathbf{N}_v'$ for realistic geometry reconstruction:
\begin{align}
L_{A_{2D}} = & \mathbb{E}_{\mathbf{N}_v}(\log \mathrm{Disc}(\mathbf{N_v})) + \nonumber \\
 & \mathbb{E}_{\mathbf{N}_v'}(\log(1-\mathrm{Disc}(\mathbf{N}_v'))),
\end{align}
where $\mathrm{Disc}(\cdot)$ denotes the CNN-based discriminator network.
The final VQ-VAE objective $\mathcal{L}_{VQVAE}$ combines both 3D and 2D losses:
\begin{equation}
\mathcal{L}_{VQVAE} = \mathcal{L}_R + \lambda \mathcal{L}_C + (1 - \lambda) \mathcal{L}_{CB} + \gamma_R \mathcal{L}_{R_{2D}} + \gamma_A L_{A_{2D}},
\end{equation}
where $\lambda$, $\gamma_R$, and $\gamma_A$ are hyperparameters balancing the different losses. Our experiments (\cref{sec:ablation}) demonstrate that this 2D supervision enhances reconstruction quality, especially in capturing finer geometric details, leading to an improved encoding $\mathbf{z}$ of the input TSDFs.

\subsection{Latent Diffusion Model for 3D Shapes}
\label{section:method_latent_diff}
Utilizing our pre-trained VQ-VAE, we remap TSDFs into a low-dimensional latent space for efficiently training our diffusion model. Fundamentally, diffusion models \cite{ho2020ddpm,rombach2022latentdiffusionmodels} learn a data distribution $p(\mathbf{z})$ by denoising a variable $\mathbf{\mathbf{z}_T}$, starting from a normal distribution. Our goal is to model this reverse diffusion process with a deep neural network with parameters $\mathbf{\theta}$.

The forward diffusion process is a fixed Markov chain, adding Gaussian noise to a latent sample $\mathbf{z}$ over time steps $t \in {1,...,T}$, following a variance schedule $\beta_1, \beta_2, ..., \beta_T$~\cite{ho2020ddpm}.
\begin{align}
& q(\mathbf{z}_{1:T} | \mathbf{z}_0) := \prod_{t=1}^T q(\mathbf{z}_t | \mathbf{z}_{t-1}) \text{,}  \nonumber \\
 & q(\mathbf{z}_t|\mathbf{z}_{t-1}) :=  \mathcal{N} (\sqrt{1-\beta_t}\mathbf{z}_{t-1}, \beta_t \mathbf{I}).
    \label{eqn:forward_process}
\end{align}
Importantly, the forward process allows to sample $\mathbf{z}_t$ at an arbitrary timestep $t$ in closed form: 
\begin{equation}
    q(\mathbf{z}_t|\mathbf{z}_0) = \mathcal{N}(\sqrt{\bar{\alpha}_t}\mathbf{z}_0, (1-\bar{\alpha}_t)\mathbf{I}) \text{,}
    \label{eqn:forward_sampling}
\end{equation}
where $\alpha_t := 1-\beta_t$ and $\bar{\alpha}_t := \prod_{s=1}^t \alpha_s$. This formulation makes it possible to sample arbitrary $\mathbf{z}_t$ without having to process $\mathbf{z}_0 ..., \mathbf{z}_{t-1}$.

The reverse process is defined as a learned Markov chain with learned Gaussian transitions starting at  $p(\mathbf{z}_T)=\mathcal{N}(\mathbf{z}_T;\mathbf{0},\mathbf{I})$ as follows \cite{ho2020ddpm}:
\begin{align}
     &p_{\theta}(\mathbf{z}_{0:T}):=p(\mathbf{z}_T)\prod_{t=1}^T p_{\theta}(\mathbf{z}_{t-1}|\mathbf{z}_t, \mathbf{c}, \mathbf{I})  \text{,} \nonumber \\
     & p_{\theta} (\mathbf{z}_{t-1}|\mathbf{z}_t) := \mathcal{N}(\bm{\mu}_{\theta}(\mathbf{z}_t,t, \mathbf{c}, \mathbf{I}), \bm{\Sigma}_{\theta}(\mathbf{z}_t,t, \mathbf{c}, \mathbf{I})) \text{,}
     \label{eqn:backward_process}
\end{align}

where $\mathbf{c}$ and $\mathbf{I}$ denote the partial TSDF and RGB image, respectively, and $\bm{\mu}_{\theta}$ and $\bm{\Sigma}_{\theta}$ denote the mean and covariance predicted by our network in latent space. Learning the reverse diffusion process involves estimating $\bm{\mu}_{\theta}$ and $\bm{\Sigma}_{\theta}$. However, it has been shown experimentally that setting $\bm{\Sigma}_{\theta}(\mathbf{z}_t,t,\mathbf{c}, \mathbf{I} ) = \sigma_t^2 \mathbf{I}$ with $\sigma_t^2 = \tilde{\beta}_t = (1-\bar{\alpha}_{t-1})/({1-\bar{\alpha}_t}) \beta_t$, $\alpha_t := 1-\beta_t$, and $\bar{\alpha}_t := \prod_{s=1}^t \alpha_s$ provides good results, while removing the need to learn the covariance.

We use the reparametrization and objective $\mathcal{L}_{DDPM}$ proposed by \cite{ho2020ddpm} (see \cref{eqn:simple_objective}) and focus on predicting the noise $\bm{\epsilon}_{\theta} (\mathbf{z}_t,t)$ added during the forward process.
\begin{equation}
\mathcal{L}_{DDPM} := \mathbb{E}_{t,\mathbf{z},\bm{\epsilon}} \left[\| \bm{\epsilon} - \bm{\epsilon}_{\theta} ( \mathbf{z}_t, t, \mathbf{c}, \mathbf{I}) \|_2^2 \right].
\label{eqn:simple_objective}
\end{equation}

At training time, we take a complete TSDF voxel grid $\mathbf{X}$ \ and compute a latent code $\mathbf{z}_0$ using our VQ-VAE's encoder $\mathcal{E}$. Then, we sample a noised latent code $\mathbf{z}_t$ at time $t$. The denoising diffusion model predicts the noise $\bm{\epsilon}_{\theta} ( \mathbf{z}_t, t)$ and is trained with the objective shown in \cref{eqn:simple_objective}.
In contrast, at inference time, we sample $\mathbf{\hat{z}}$ by gradually denoising a noise variable $\mathbf{z}_T$ sampled from the standard normal distribution $\mathcal{N}(\mathbf{z}_T;\mathbf{0},\mathbf{I})$. We then use our pretrained decoder $\mathcal{D}$ and codebook $\mathcal{Z}$ to map the denoised code $\mathbf{\hat{z}}$ back to a predicted complete TSDF voxel grid $\mathbf{\hat{X}}$. Our denoising diffusion model $\bm{\epsilon}_{\theta} ( \mathbf{z}_t, t)$ is implemented as a time-conditional 3D U-Net ~\cite{ho2020ddpm,rombach2022latentdiffusionmodels}.

\subsection{Conditioning Mechanisms}
\label{section:method_conditioning}
We propose the use of two complementary conditioning mechanisms, as illustrated in \cref{fig:overview}: cross-attention on CLIP features \cite{radford2021_clip} extracted from an object image, and spatially consistent feature aggregation from the TSDF voxel grid of a partial scan.

\textbf{Conditioning on RGB Images}
\label{section:crossattn_conditioning}
Diffusion models can model conditional distributions of the form $p(\mathbf{z}|\mathbf{y})$, which allows us to implement a conditional denoising autoencoder $\epsilon_{\theta}(\mathbf{z}_t,t,\mathbf{y})$, where $\mathbf{y}$ denotes a conditioning input.
Following the trend of latent diffusion models \cite{rombach2022latentdiffusionmodels}, we extend the U-Net backbone with a cross-attention mechanism. This enables learning the distribution of latent codes $p(\mathbf{z}|\mathcal{E}_I(\mathbf{I}))$ conditioned on pre-trained CLIP encoder features of an RGB image $\mathbf{I}$ of the 3D object, rendered from the viewpoint corresponding to the partial scan.
The CLIP encoder $\mathcal{E}_I$ projects the RGB image $\mathbf{I} \in \mathbb{R}^{3\times 256 \times 256}$ to an intermediate representation $\mathcal{E}_I(\mathbf{I}) \in \mathbb{R}^{M \times d_{\mathcal{E}_I}}$, where $d_{\mathcal{E}_I} = 768$ is the feature dimension and $M=50$ corresponds to the flattened spatial dimension of the CLIP encoding. This representation is then injected into the intermediate layers of the time-conditional U-Net via cross-attention layers.

\textbf{Conditioning on Partial Scans}
\label{section:control_conditioning}
To improve control over spatial structure and overcome the limitations of image-only conditioning, we introduce a second mechanism inspired by ControlNet and DiffComplete~\cite{zhang2023controlnet,chu2023diffcomplete}. Instead of artificially corrupting the ground-truth TSDF map, we employ a mapping pipeline to obtain realistic partial reconstructions from a single viewpoint \cite{nie2020total3dunderstanding,curless1996volumetric_fusion}. Details of the partial TSDF preprocessing are provided in the supplementary.
The corrupted ground-truth latent code $\mathbf{z}_t$ and the partial shape $\mathbf{c}$ are encoded into multi-resolution feature maps of the same size, $\mathcal{E}_{\epsilon}(\mathbf{z}_t)$ and $\mathcal{E}_c(\mathbf{c})$, respectively. During the forward pass of the diffusion model, $\mathcal{E}_{\epsilon}(\mathbf{z}_t)$ is propagated through the U-Net, while the control branch fuses the features as $\mathbf{f} = \mathcal{E}_{\epsilon}(\mathbf{z}_t) + \mathcal{E}_c(\mathbf{c})$ and forwards them into deeper layers.
The control branch connects to the diffusion model via $1 \times 1$ convolutional layers to ensure consistent feature map dimensions for aggregation. The features entering the $i$-th decoder block of the diffusion model are expressed as:
\begin{equation}
d^{(i)} = [\mathcal{D}_z^{(i-1)}(\mathbf{z}_t), F_z^{(i)}(\mathbf{z}_t)+\phi^{(i)}(F_c^{(i)}(\mathbf{f}))] \text{,}
\end{equation}
where $\phi^{(i)}$ is the $1 \times 1$ convolutional projection layer, $F_z^{(i)}(\mathbf{z}_t)$ are the encoder features of the diffusion model's $i$-th encoder block passed via residual connections to the decoder block $\mathcal{D}_z^{(i-1)}$, and $F_c^{(i)}(\mathbf{f})$ are the features from the control branch's $i$-th encoder block. Here, $[\cdot,\cdot]$ denotes concatenation.
This design ensures spatially consistent integration of partial scan features into the reverse diffusion process. Combining both conditioning approaches allows the model to leverage global shape cues from images while incorporating detailed 3D information from partial scans, resulting in realistic and high-fidelity shape completion.

%% file: chapters/04_experiments.tex
\input{figures/3depn_results}
\section{Experiments}\label{section:results}
We evaluate our method in the task of shape completion on two large-scale shape completion benchmarks.
The 3D-EPN~\cite{dai2017_3depn} dataset contains 25,590 object instances of eight classes from ShapeNet~\cite{chang2015shapenet}. For each instance, several partial scans of varying completeness are provided as $32^3$ TSDF grids, while complete shapes are represented as $32^3$ or $64^3$ truncated unsigned distance functions. 
The PatchComplete~\cite{rao2022patchcomplete} benchmark emphasizes completing objects from unseen categories. It contains 5,000 partial/complete pairs from the synthetic ShapeNet~\cite{chang2015shapenet} and 11,000 partial/complete pairs from objects in the scenes of the real-world ScanNet dataset~\cite{dai2017scannet}.

We first train a VQ-VAE for 150,000 steps. Next, we train our conditional shape completion network (diffusion model and control branch) using the same data split for other 200,000 steps, while the CLIP encoder remains frozen with pre-trained weights. In both cases we use the Adam optimizer \cite{zhang2018adam} with a learning rate of $1 \times 10^{-4}$ and $2.5 \times 10^{-5}$, respectively. 
We set the batch size to $8$ and trained both the VQ-VAE and the shape completion network on a single NVIDIA A40 GPU with 48~GB of VRAM.
While the VQ-VAE can be trained in 48 hours, training the diffusion model takes about 120 hours. On the other hand.
At inference, we use the DDIM formulation~\cite{song2020denoising_ddim}, requiring only 100 denoising steps, which take approximately 15 seconds per sample.

\subsection{Quantitative Evaluation}
As commonly used in the field, we evaluate the $l_1$ error across all voxels for known object classes as well as the chamfer distance (CD), and the intersection over union (IoU) between predicted and ground-truth shapes for unknown classes. 
\begin{figure}[ht]
    \centering 
    \footnotesize
    \begin{overpic}[width=\linewidth, trim=0 0 0 -300, clip]{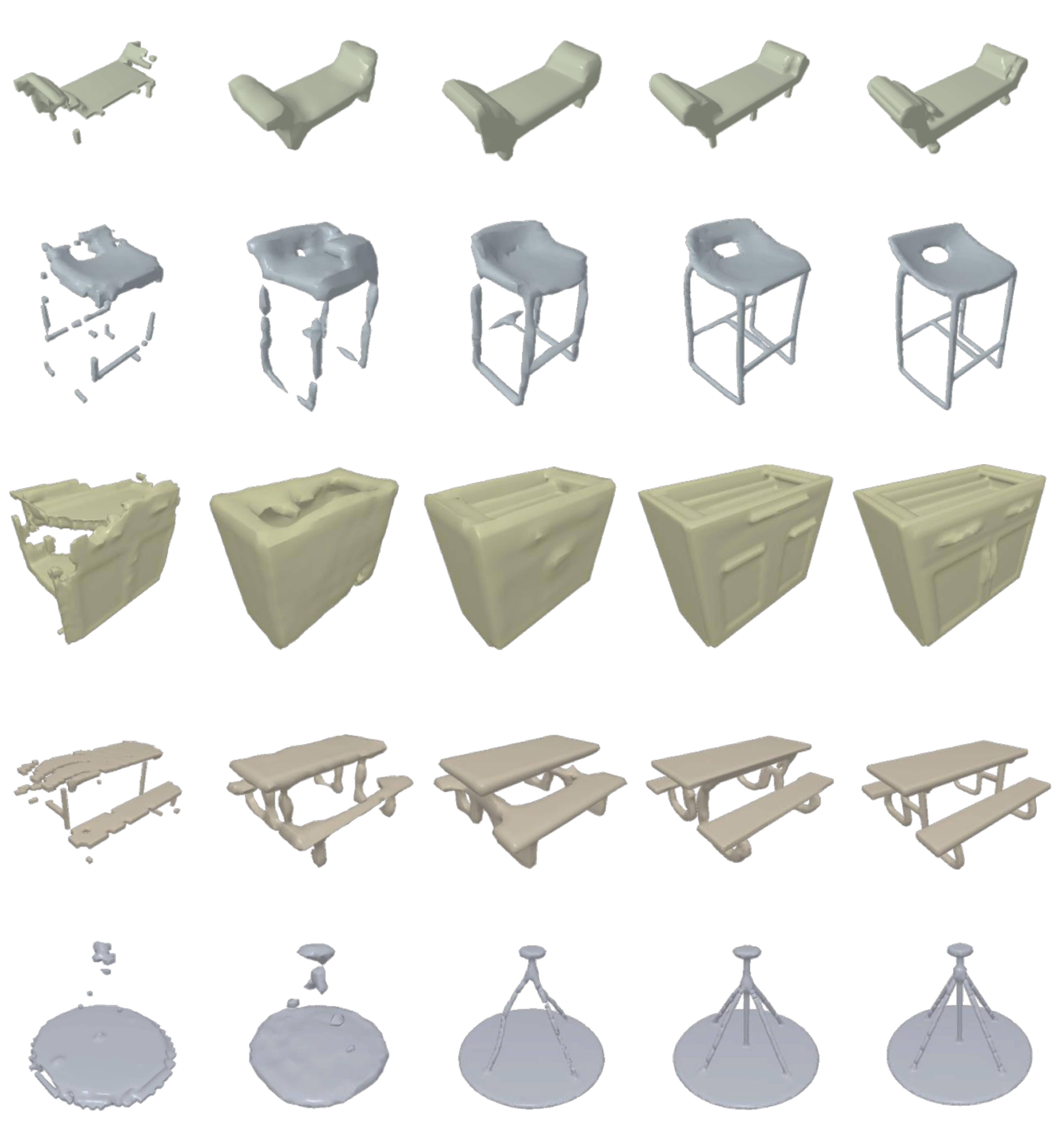}
        \put(8, 98){\makebox[0pt]{Input}}
        \put(26, 98){\makebox[0pt]{3D-EPN~\cite{dai2017_3depn}}}
        \put(45, 98){\makebox[0pt]{AutoSDF~\cite{mittal2022autosdf}}}
        \put(60, 98){\makebox[0pt]{Ours}}
        \put(79, 98){\makebox[0pt]{GT}}
    \end{overpic}
    \caption{Qualitative results on the 3D-EPN~\cite{dai2017_3depn} test set. DiffComplete~\cite{chu2023diffcomplete} results are excluded due to unavailable open-source weights. Leveraging improved model scalability, our method reconstructs finer geometric details while maintaining faithful to the input shapes.}
    \label{fig:results_3dpn}
    \vspace{-0.4cm}
\end{figure}

\textbf{3D-EPN Benchmark -- Known Object Classes}
Our method achieves superior performance compared to state-of-the-art approaches. On the standard $32^3$ resolution benchmark, our class-agnostic model performs on par with the class-specific DiffComplete~\cite{chu2023diffcomplete}, while widely surpassing all other class-agnostic baselines by $47\%$. This demonstrates strong generalization across categories without relying on class-specific supervision. Moreover, when trained in a class-specific setting, our model outperforms DiffComplete by a $12\%$ reduction in $l_1$ error while being $30\%$ more memory-efficient, highlighting the effectiveness of our latent formulation and multimodal conditioning. Importantly, unlike prior methods limited to $32^3$ due to little or no compression, our approach scales to $64^3$ TSDF voxel grids while remaining efficient, enabling higher-fidelity completions with finer geometric details.
As shown in Tab.~\ref{table:ablations_conditioning}, we further analyze the impact of different conditioning mechanisms. Notably, even when conditioned only on the partial TSDF input, our class-agnostic model still outperforms the state-of-the-art by a $19\%$ reduction in $l_1$ error, underlining the strength of our latent diffusion framework. Incorporating RGB guidance further enhances reconstruction quality, as illustrated by the qualitative results in \cref{fig:results_3depn1}, where our model excels at recovering realistic and detailed shapes from incomplete observations.

\input{figures/patchcomplete_results}
\begin{figure}[ht]
    \centering 
    \footnotesize
    \begin{overpic}[width=\linewidth, trim=0 0 0 -400, clip]{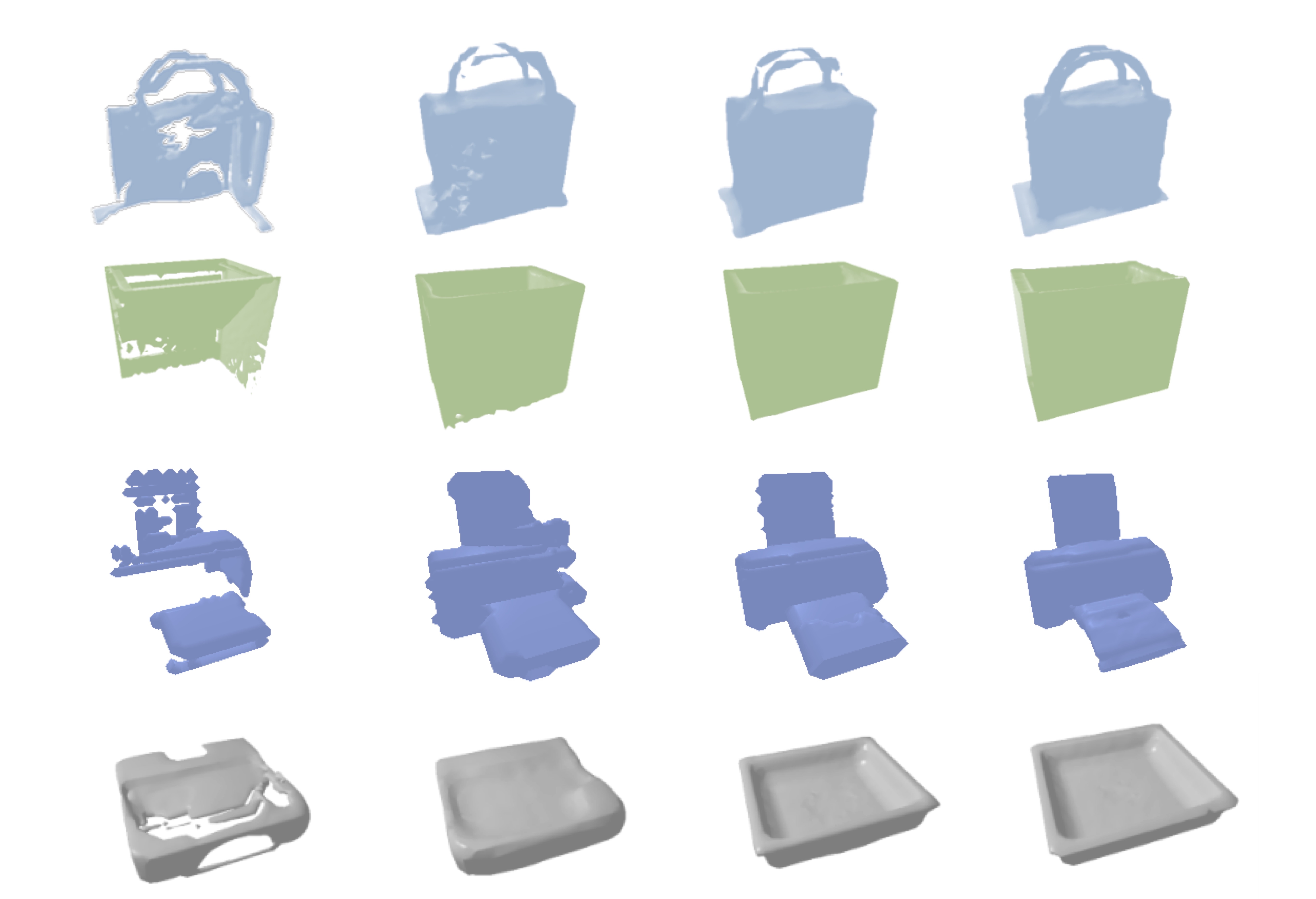}
        \put(14, 70){\makebox[0pt]{Input}}
        \put(40, 70){\makebox[0pt]{PatchComplete~\cite{rao2022patchcomplete}}}
        \put(60, 70){\makebox[0pt]{Ours}}
        \put(83, 70){\makebox[0pt]{GT}}
    \end{overpic}
    \caption{Qualitative results on the PatchComplete~\cite{rao2022patchcomplete} benchmark for unseen categories. Note the generally smoother surfaces and sharper edges produced by our model.}
    \label{fig:results_patchcomplete}
    \vspace{-0.2cm}
\end{figure}

\textbf{PatchComplete Benchmark -- Unseen Object Classes}
For object categories not observed during training, our method outperforms competitors in almost all specific categories, demonstrating an improvement for both synthetic and real-world data (see \cref{table:patchcomplete_results}). \cref{fig:results_patchcomplete} shows some of the predictions performed by our method on this benchmark, showcasing its superior ability to complete shapes from unseen categories with inputs of various levels of completeness.

\begin{figure}[ht]
    \raggedleft 
    \footnotesize
    \begin{overpic}[width=0.85\linewidth, trim=0 0 0 -800, clip]{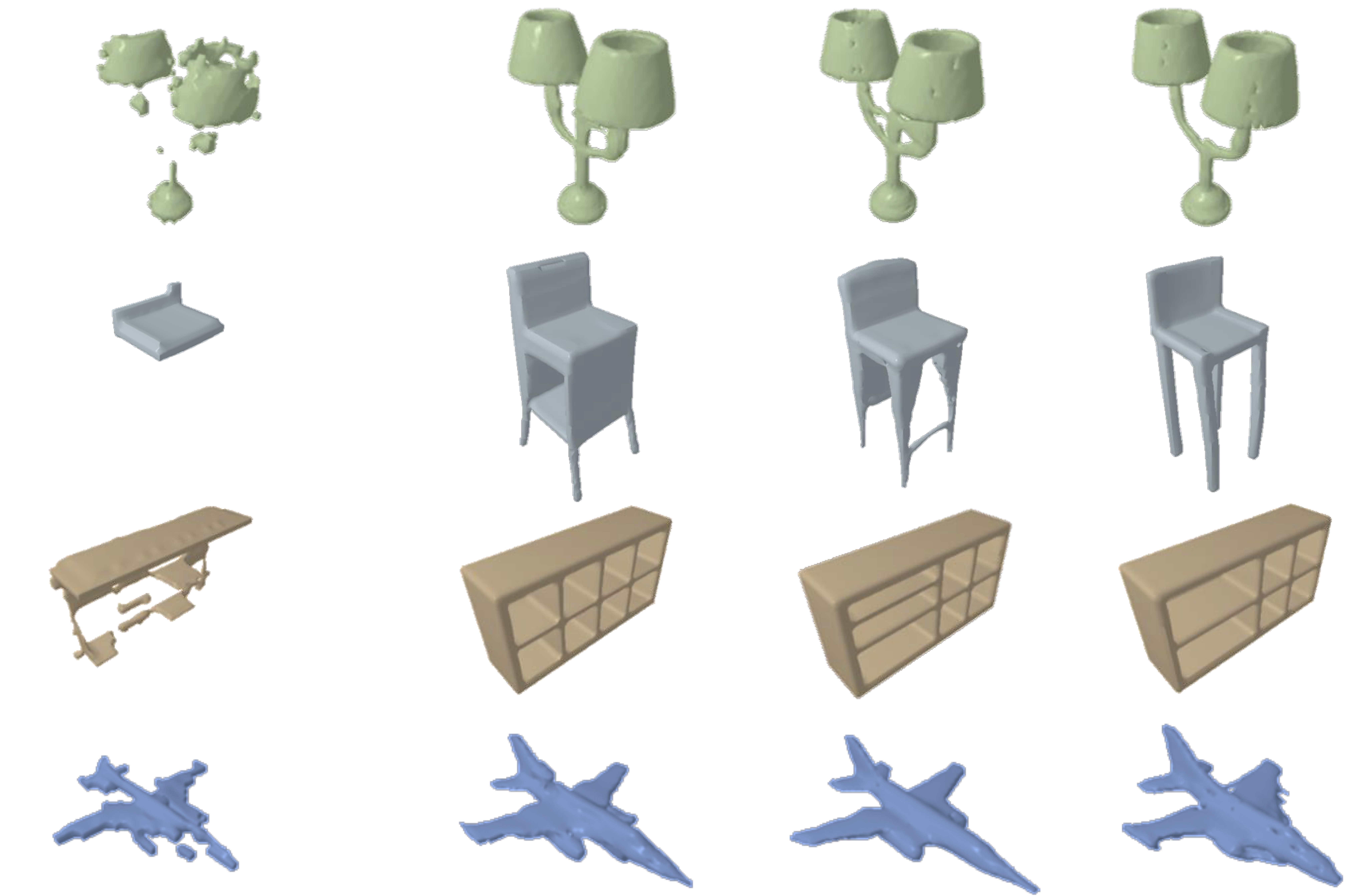}
        \put(13, 70){\makebox[0pt]{Input}}
        \put(65, 70){\makebox[0pt]{Diverse Samples}}
        \put(-10, 26){\makebox[0pt]{\rotatebox{90}{Ours}}}
    \end{overpic}
    \begin{overpic}[width=0.85\linewidth, trim=0 0 0 -200, clip]{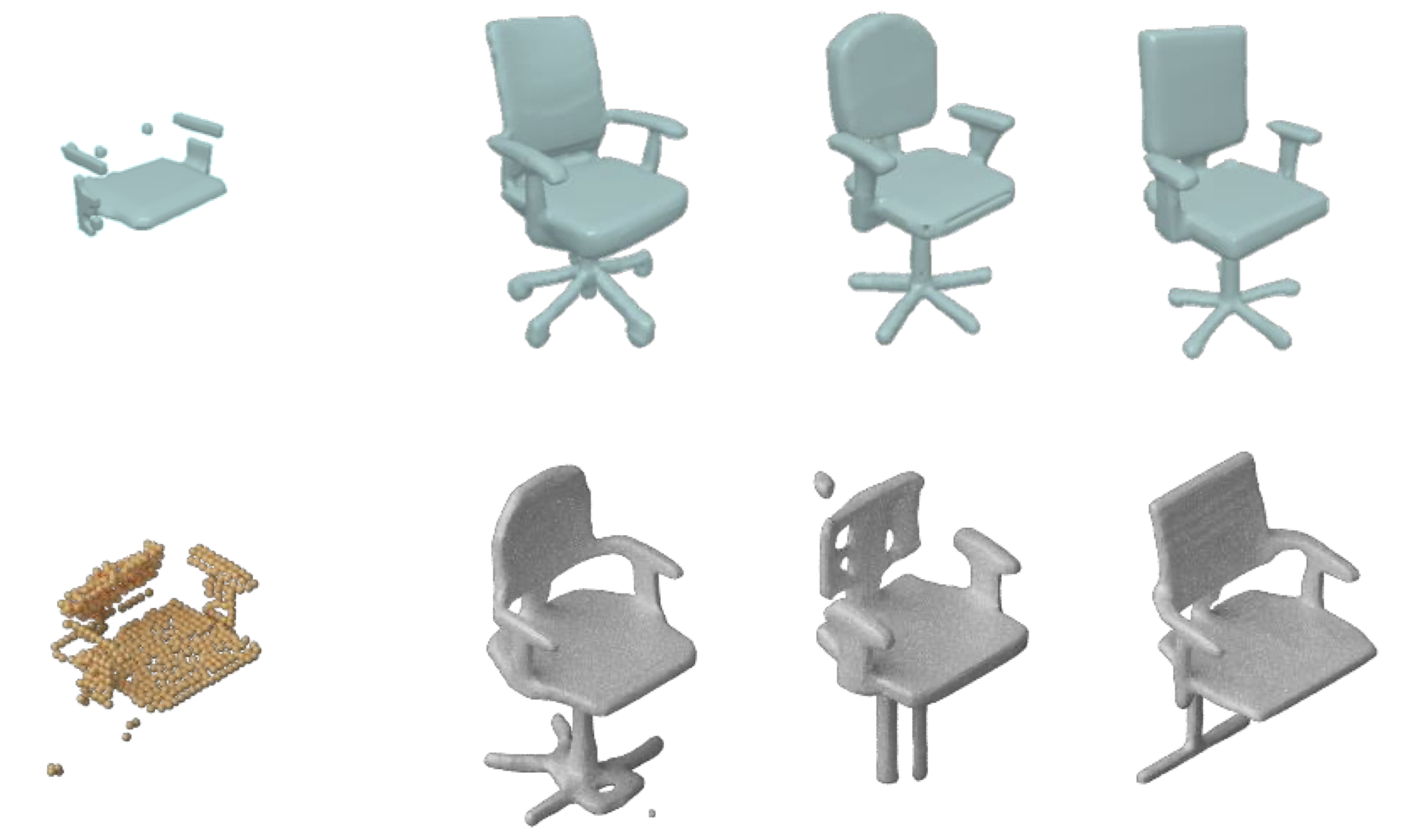}
        \put(-10, 2){\makebox[0pt]{\rotatebox{90}{ShapeFormer~\cite{yan2022shapeformer}\quad\quad\quad Ours}}}
    \end{overpic}
    \caption{Our approach produces diverse but realistic and detailed multifaceted shape completions for the same input partial scans, for both real-world (top) and synthetic (bottom) examples.}
    \label{fig:multimodality}
    \vspace{-0.4cm}
\end{figure}

It is noteworthy to highlight the probabilistic nature of our method, which inherently allows for multiple potential shape completions from the same input, as depicted in \cref{fig:multimodality}. This characteristic is particularly advantageous in the context of shape completion, a problem intrinsically under-constrained and characterized by a multitude of feasible solutions. Consequently, it is beneficial for a method to not only generate realistic, high-fidelity outcomes but also to offer a diversity of results that encapsulate the range of possible solutions. Our approach successfully achieves this balance, showcasing its versatility in handling the complexity of the shape completion task.

\input{figures/shapenet_results}
\textbf{Comparison to Non-TSDF-Based Approaches}
While our method and other previously discussed approaches operate on TSDF grids, the methods compared in \cref{tab:sdfusion} use different 3D representations: SDFusion~\cite{cheng2023sdfusion} relies on signed distance functions (SDFs), and ShapeFormer~\cite{yan2022shapeformer} operates on point clouds. For a fair comparison with SDFusion~\cite{cheng2023sdfusion}, we followed their shape completion strategy by generating partial inputs through masking significant portions of the ground-truth shapes and finetuned our model for an additional 50,000 steps.
The results in \cref{tab:sdfusion} show that our TSDF-based approach substantially outperforms both SDFusion~\cite{cheng2023sdfusion} and ShapeFormer~\cite{yan2022shapeformer} across all metrics. Qualitative examples in  \cref{fig:multimodality} further illustrate that our method produces more plausible shapes with higher structural fidelity and greater shape diversity. Additional qualitative comparisons can be found in the supplementary.

\input{figures/ablation_vqvae_2d}

\input{figures/ablation_conditioning}

\subsection{Ablation Studies}\label{sec:ablation}
\begin{figure}[ht]
    \centering 
    \footnotesize
    \begin{overpic}[width=0.7\linewidth, trim=0 0 0 -300, clip]{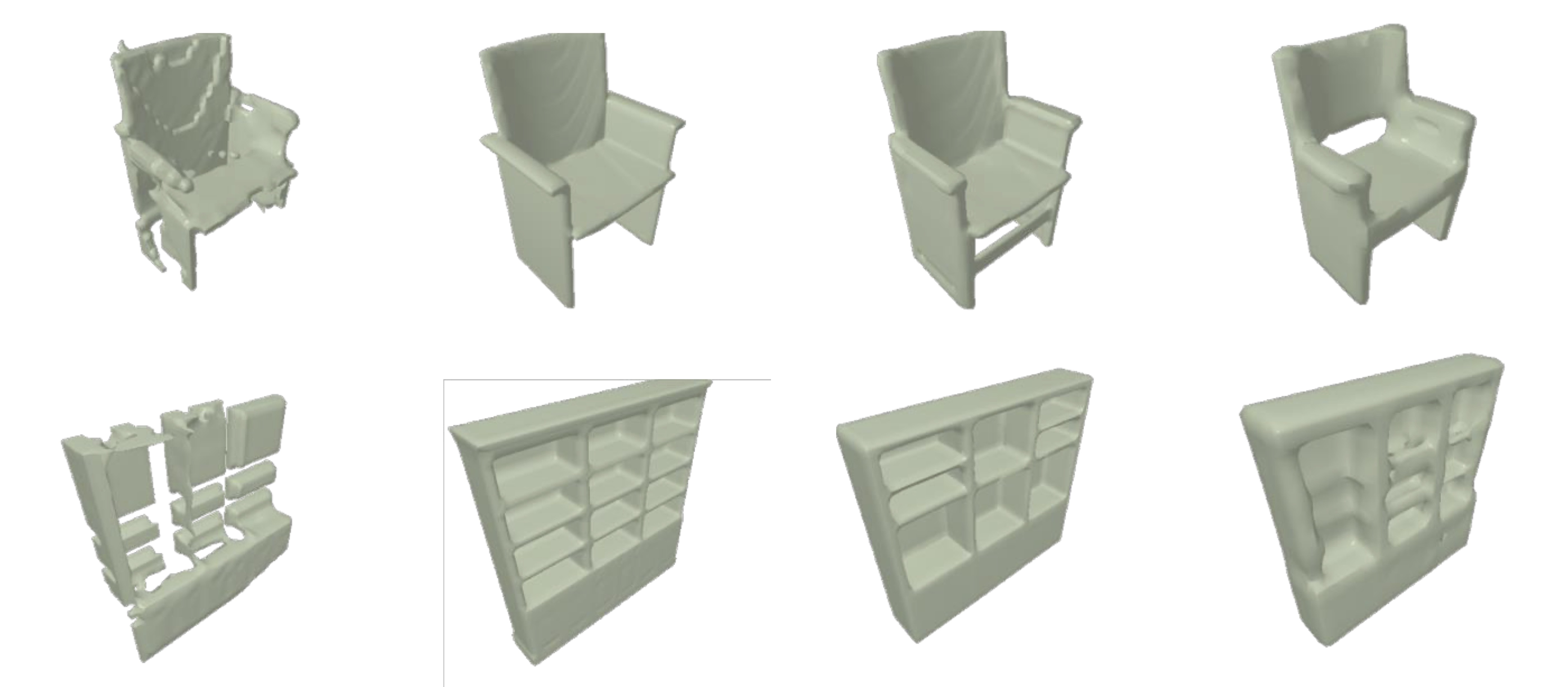}
        \put(11, 48){\makebox[0pt]{Input}}
        \put(36, 48){\makebox[0pt]{GT}}
        \put(60, 48){\makebox[0pt]{64}}
        \put(85, 48){\makebox[0pt]{32}}
    \end{overpic}
    \caption{Completions for various resolutions on ShapeNet~\cite{chang2015shapenet}.}
    \label{fig:different_resolutions}
    \vspace{-0.3cm}
\end{figure}

\textbf{Effect of Increasing Resolutions}
\cref{fig:different_resolutions} illustrates shape completions at different output resolutions. While our low-resolution model achieves state-of-the-art performance, the reduced grid size inevitably oversmoothes complex geometries, such as shelves and chairs. In contrast, our higher-resolution outputs at $64^3$ preserve significantly finer structural details.

\textbf{Influence of 2D Loss on VQ-VAE Performance}
We extend the original VQ-VAE \cite{van2017vqvae1} by incorporating supplementary 2D losses that operate on 2D renderings of the reconstructed TSDFs. \cref{table:ablations_vqvae_2d} demonstrates the beneficial impact of these added 2D objectives on the autoencoder's shape reconstructions, showing a substantial enhancement in reconstruction quality, with a $54\%$ reduction in $l_1$ error and a $6\%$ increase in IoU. 
Qualitative illustrations of the efficacy of integrating a 2D objective to create high-fidelity reconstructions are shown in the appendix.
As shown, our training strategy results in reconstructions that more accurately reflect the original input shapes. Such high-fidelity reconstruction is essential, given that the efficient encoding of the TSDF space into a compact latent representation is critical for the success of our conditional shape completion diffusion model.

\textbf{Effects of Different Conditioning Types}
\cref{table:ablations_conditioning} evaluates the contribution of our conditioning mechanisms on the 3D-EPN benchmark across three scenarios: image-based conditioning alone, partial shape-based conditioning alone, and their combination. While each conditioning mechanism provides unique benefits, their concurrent use leads to a further $15\%$ reduction in $l_1$ error compared to using partial shape conditioning alone. This highlights that the combination effectively leverages complementary information and enables a single model to perform competitive shape completion across multiple object categories.

%% file: figures/3depn_results.tex
\begin{table*}[ht!]
  \caption[Shape completion results on the 3D-EPN benchmark]{Quantitative comparison on the 3D-EPN benchmark \cite{dai2017_3depn}. Our class-specific model consistently outperforms all state-of-the-art methods in reconstruction accuracy, while it still achieves competitive results when generalizing over all classes.}
  \label{table:3depn_results}
  \centering
  \begin{tabular}{l | c c c c c c c}
    \toprule
    \( l_1 \) error \( \downarrow \) & 3D-EPN & AutoSDF & PatchComplete & DiffComplete & \textbf{Ours} & \textbf{Ours} & \textbf{Ours} \\
    & \cite{dai2017_3depn} & \cite{mittal2022autosdf} & \cite{rao2022patchcomplete} & \cite{chu2023diffcomplete} & &  \\
    \midrule
    Resolution & $32^3$ & $32^3$ & $32^3$ & $32^3$ & $32^3$ & $64^3$ & $64^3$\\
    Class-specific & x & x & x &\checkmark & x & x & \checkmark \\ 
    \midrule
    Chair & 0.418 & 0.201 & 0.134 & \textbf{0.070}& 0.082 & 0.086 & 0.078\\
    Table & 0.377 & 0.258 & 0.095 & 0.073& 0.074 & 0.074 & \textbf{0.072}\\
    Sofa & 0.392 & 0.226 & 0.084 & 0.061 & 0.070 & 0.075 & \textbf{0.055}\\
    Lamp & 0.388 & 0.275 & 0.087 & 0.059 & 0.047 & 0.048 & \textbf{0.037}\\
    Plane & 0.421 & 0.184 & 0.061 & 0.015& 0.015 & 0.017 & \textbf{0.014}\\
    Car & 0.259 & 0.187 & 0.053 & \textbf{0.025}& 0.032 & 0.041 & 0.033\\
    Cabinet & 0.381 & 0.248 & 0.134 & 0.086 & 0.095 & 0.091 & \textbf{0.065}\\
    Watercraft & 0.356 & 0.157 & 0.058 & 0.031& 0.028 & 0.028 & \textbf{0.024}\\
    \midrule
    Avg. & 0.374 & 0.217 & 0.088 & 0.053& 0.055 & 0.058 & \textbf{0.047}\\
    \bottomrule
  \end{tabular}
\end{table*}

%% file: figures/patchcomplete_results.tex
\begin{table*}[ht]
  \caption{Quantitative comparison for shape completion results for unknown categories on the PatchComplete benchmark \protect\cite{rao2022patchcomplete} at a $32^3$ resolution. Top: results for synthetic shapes from the ShapeNet dataset \cite{chang2015shapenet}. Bottom: results for real objects from the Scannet dataset \cite{dai2017scannet}}
  \label{table:patchcomplete_results}
  \centering
  \begin{tabular}{l | c c c c c}
    \toprule
    CD$\downarrow$/IoU$\uparrow$ & 3D-EPN & Auto-SDF & PatchComplete & DiffComplete & \textbf{Ours} \\
    & \cite{dai2017_3depn} & \cite{mittal2022autosdf} & \cite{rao2022patchcomplete} & \cite{chu2023diffcomplete} & \\
    \midrule
    Bag & 5.01 / 73.8 & 5.81 / 56.3 & 3.94 / 77.6 & 3.86 / 78.3 & \textbf{3.79} / \textbf{78.3} \\
    Lamp & 8.07 / 47.2 & 6.57 / 39.1 & \textbf{4.68} / 56.4 & 4.80 / 57.9 & 4.74 / \textbf{60.0} \\
    Bathtub & 4.21 / 57.9 & 5.17 / 41.0 & 3.78 / 66.3 & \textbf{3.52} / \textbf{68.9} & 3.67 / 65.9 \\
    Bed & 5.84 / 58.4 & 6.01 / 44.6 & 4.49 / 66.8 & \textbf{4.16} / 67.1 & 4.40 / \textbf{67.1} \\
    Basket & 7.90 / 54.0 & 6.70 / 39.8 & 5.15 / 61.0 & 4.94 / 65.5 & \textbf{4.89} / \textbf{68.5} \\
    Printer & 5.15 / 73.6 & 5.82 / 49.9 & 4.63 / \textbf{77.6} & 4.40 / 76.8 & \textbf{4.36} / 76.8 \\
    Laptop & 3.90 / 62.0 & 4.81 / 51.1 & 3.77 / 63.8 & 3.52 / 67.4 & \textbf{3.41} / \textbf{68.4} \\
    Bench & 4.54 / 48.3 & 4.31 / 39.5 & 3.70 / 53.9 & 3.56 / 58.2 & \textbf{3.39} / \textbf{61.1} \\
    \midrule
    Avg. & 5.58 / 59.4 & 5.86 / 45.2 & 4.27 / 65.4 & 4.10 / 67.5 & \textbf{4.08} / \textbf{68.2} \\
    \bottomrule
  \end{tabular}
  \begin{tabular}{l | c c c c c}
    \toprule
    CD$\downarrow$/IoU$\uparrow$ & 3D-EPN & Auto-SDF & PatchComplete & DiffComplete & \textbf{Ours} \\
    & \cite{dai2017_3depn} & \cite{mittal2022autosdf} & \cite{rao2022patchcomplete} & \cite{chu2023diffcomplete} & \\
    \midrule
    Bag & 8.83 / 53.7 & 9.30 / 48.7 & 8.23 / \textbf{58.3} & \textbf{7.05} / 48.5 & 7.41 / 50.0 \\
    Lamp & 14.3 / 20.7 & 11.2 / 24.4 & 9.42 / 28.4 & 6.84 / 30.5 & \textbf{6.39} / \textbf{33.2} \\
    Bathtub & 7.56 / 41.0 & 7.84 / 36.6 & \textbf{6.77} / 48.0 & 8.22 / 48.5 & 8.09 / \textbf{48.5} \\
    Bed & 7.76 / 47.8 & 7.91 / 38.0 & 7.24 / 48.4 & 7.20 / 46.6 & \textbf{6.91} / \textbf{48.6} \\
    Basket & 7.74 / 36.5 & 7.54 / 36.1 & 6.60 / 45.5 & 7.42 / 59.2 & \textbf{6.38} / \textbf{62.2} \\
    Printer & 8.36 / 63.0 & 9.66 / 49.9 & 6.84 / 70.5 & \textbf{6.36} / \textbf{74.5} & 7.10 / 69.1 \\
    \midrule
    Avg. & 9.09 / 44.0 & 8.90 / 38.9 & 7.52 / 49.5 & 7.18 / 51.3 & \textbf{7.04} / \textbf{51.9} \\
    \bottomrule
  \end{tabular}
\end{table*}

%% file: figures/shapenet_results.tex
\begin{table}[h]
\centering
\begin{tabular}{c | c c c}
\toprule
Metric & SDFusion~\cite{cheng2023sdfusion} & ShapeFormer~\cite{yan2022shapeformer} & Ours \\
\midrule
$l_1$ error $\downarrow$ & 0.086 & 0.097 & \textbf{0.068} \\
CD $\downarrow$& 0.058 & 0.073 & \textbf{0.025} \\
IoU $\uparrow$& 0.38 & 0.31 & \textbf{0.55} \\
\bottomrule
\end{tabular}
\caption{Quantitative comparison between SDFusion\cite{cheng2023sdfusion}, ShapeFormer \cite{yan2022shapeformer}, and our method on the ShapeNet dataset~\cite{chang2015shapenet}.}
\label{tab:sdfusion}
\end{table}

%% file: figures/ablation_vqvae_2d.tex
\begin{table}[ht]
  \caption{Effect of 2D losses on the VQ-VAE reconstructions.}
  \label{table:ablations_vqvae_2d}
  \centering
  \begin{tabular}{c | c c}
    \toprule
    2D loss weight $(\gamma_A)$ & $l_1$ error $\downarrow$ & IoU $\uparrow$  \\
    \midrule
    0.0 & 6.49$e^{-3}$  & 88.48 \\
    0.2 & 4.06$e^{-3}$  & 91.54 \\
    0.4 & 2.99$e^{-3}$ & \textbf{93.15} \\
    0.6 & 3.48$e^{-3}$  & 92.97 \\
    \bottomrule
   \end{tabular}
\end{table}

%% file: figures/ablation_conditioning.tex
\begin{table}[ht]
    \caption{Impact of the different conditioning mechanisms evaluated on the 3D-EPN dataset.}
    \label{table:ablations_conditioning}
    \centering
    \begin{tabular}{c c|c }
        \toprule
        Image-based & Partial TSDF & Avg. $l_1$ error $\downarrow$ \\
        \midrule
        \checkmark &  &  0.341\\
         & \checkmark &  0.072 \\
        \checkmark & \checkmark &  \textbf{0.058} \\
        \bottomrule
    \end{tabular}  
\end{table}

%% file: chapters/05_conclusion.tex
\section{Conclusion}
\label{section:conclusion}
In this work, we presented a novel framework for 3D shape completion that unifies latent diffusion models with both 2D and 3D conditioning in a spatially consistent manner. By jointly training a VQ-VAE with 2D and 3D supervision, we compress TSDF voxel grids into a compact discrete latent space, enabling $64^3$ resolution reconstructions while reducing GPU memory consumption by $30\%$ compared to state-of-the-art approaches.
Beyond architectural efficiency, our method advances conditional generation by combining cross-attention with ControlNet-inspired guidance, allowing flexible integration of complementary modalities. To better capture real-world challenges, we introduce a training strategy that simulates realistic partial scans directly from 3D datasets without prior assumptions on input structure. This results in a robust multimodal framework capable of class-agnostic generation while producing diverse and high-fidelity shape completions.
Comprehensive experiments on both synthetic and real-world benchmarks confirm that our approach achieves state-of-the-art performance, reducing $l_1$ reconstruction error by $12\%$ over class-specific baselines and by $47\%$ over class-agnostic models. These results highlight the advantages of our method in overcoming resolution bottlenecks and improving the realism of 3D completions.
Despite these contributions, challenges remain in extending the method to complex scenes. Future work will explore recent advances in efficient diffusion, such as non-Markovian sampling, progressive distillation~\cite{salimans2022progressive_diffusion_distillation}, and consistency models~\cite{song2023consistency}. We believe these avenues will further enhance the practicality of latent diffusion for real-world 3D applications, enabling high-quality, scalable shape completion.

%% file: chapters/X_supplementary.tex
\clearpage
\setcounter{page}{1}
\setcounter{section}{0}
\renewcommand\thesection{\Roman{section}}
\maketitlesupplementary

\section{VQ-VAE Ablations}
\begin{figure}[ht]
  \centering
  \includegraphics[width=220pt]{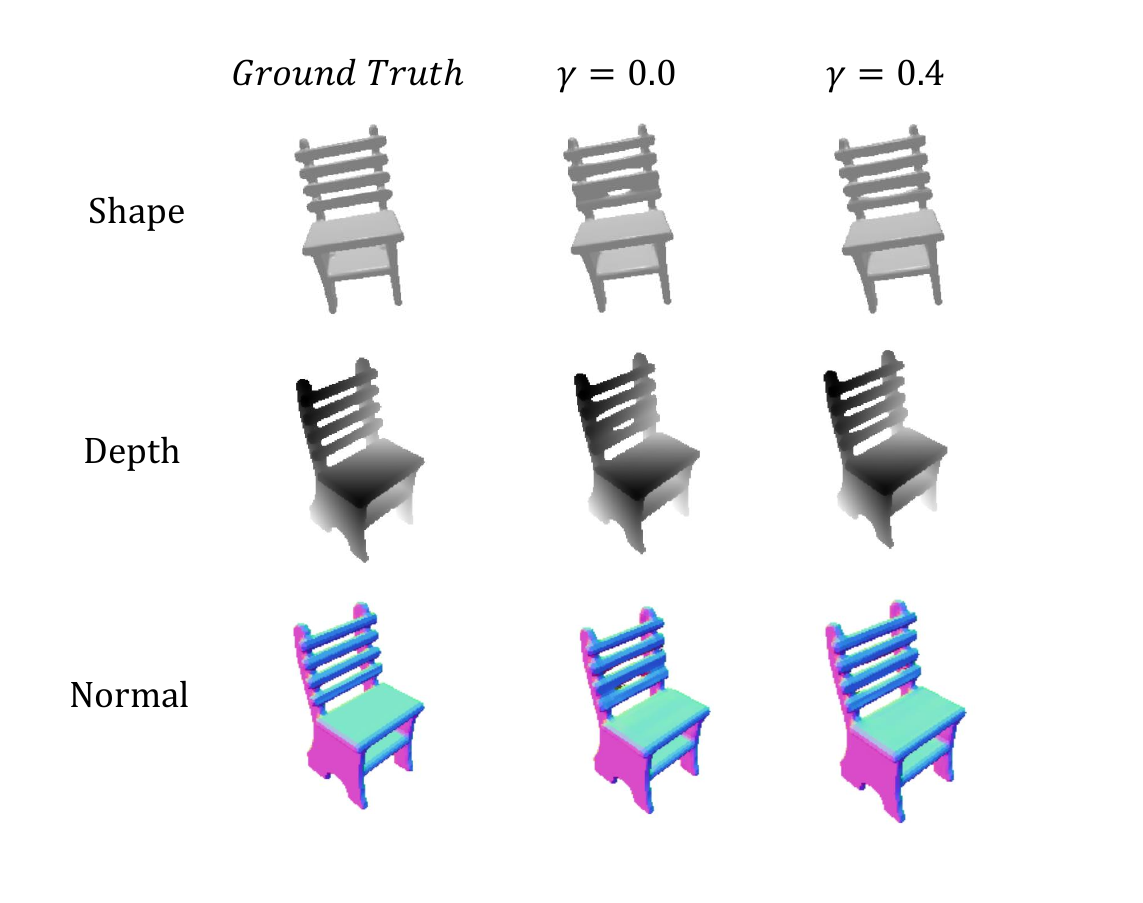}
  \caption{The introduction of 2D losses produces much cleaner reconstructions in the VQ-VAE, enabling modeling in a compact latent space while preserving crisp geometric details.}
  \label{fig:vqvae_ablation}
\end{figure}

\cref{fig:vqvae_ablation} shows a qualitative illustration of the efficacy of integrating a 2D objective to create high-fidelity reconstructions. As illustrated, the 2D losses help improve the reconstruction of fine-grained details of the object, such as the individual elements of a chair's backrest.

\section{Effect of Different Conditioning Types}
\begin{figure}[ht]
   \centering
   \includegraphics[width=260pt]{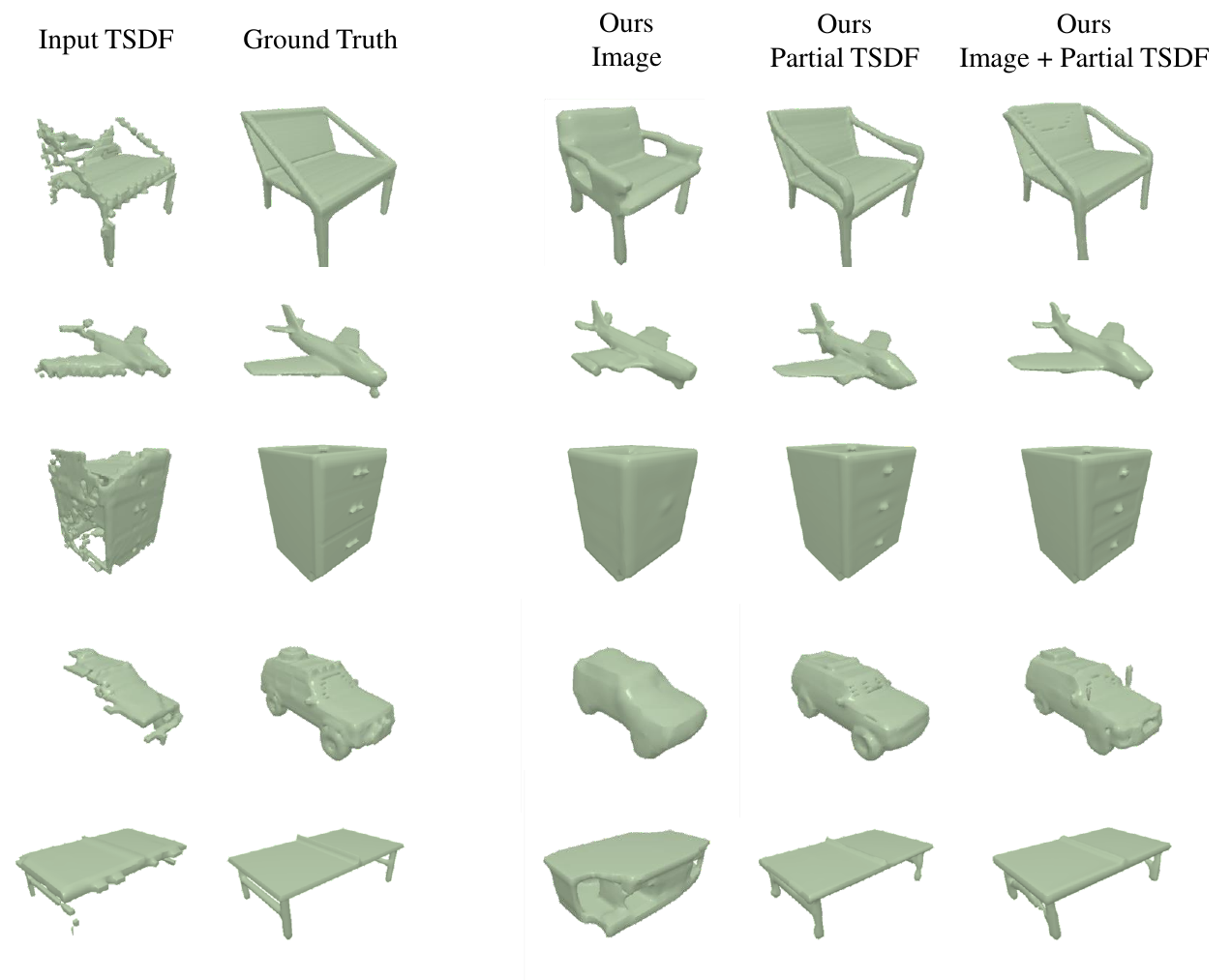}
   \caption{Comparison of conditioning modalities. Using only TSDF, the diffusion model successfully completes partial shapes but lacks fine detail. When conditioned on both TSDF and RGB, the model exploits appearance cues to produce sharper and more detailed reconstructions.}
   \label{fig:conditioning}
\end{figure}

\cref{fig:conditioning} illustrates a visual comparison of the proposed conditioning mechanisms: Image Only, Partial TSDF Only, and Multimodal Conditioning. We conducted five predictions on each input using the specified conditioning mechanism and selected the prediction with the lowest L1 error. While image-only conditioning leads to severe over-smoothing of the reconstructed object, only conditioning on the partial TSDF misses many details. The results underscore the enhanced effectiveness of utilizing both conditioning mechanisms in tandem over image-only conditioning, which proves insufficient to obtain high-quality results.

\section{Preprocessing}
\label{sec:preprocessing}
We employed the ShapeNetCore dataset~\cite{chang2015shapenet}, which contains 55 categories of common objects. For our model, we required data tuples $(\mathbf{X}, \mathbf{c}, \mathbf{I})$ including complete and partial TSDF grids and RGB images. Given that ShapeNet provides objects' meshes and textures, we converted these into the necessary TSDF voxel grids. Our process involved rendering 20 views (RGB+Depth) of each object using depth rendering techniques~\cite{NEURIPS2020_depth_rendering}, followed by volumetric fusion~\cite{curless1996volumetric_fusion} to generate the complete TSDF grid $\mathbf{X}$. For the partial TSDF grid $\mathbf{c}$, we randomly select one of these views and fill out the TSDF voxel-grid using the camera pose and depth such that observed regions of space have a positive value while unobserved regions of space have the negative truncation value.
The corresponding RGB image from this view is denoted as $\mathbf{I}$.

\section{Network Architecture}
Our diffusion model is fundamentally a U-Net \cite{ronneberger2015unet} with an encoder, a middle block, and a decoder, which is connected to the encoder through skip connections. Both encoder and decoder contain ResNetBlocks. In the case of the encoder, these are accompanied by downsampling blocks, while in the decoder, they are accompanied by up-sampling convolution layers. The encoder, decoder, and middle block include two spatial transformers each. Images are encoded using CLIP encoder \cite{radford2021_clip}, and diffusion timesteps are encoded with a 2-layer MLP using positional encoding. 
The control branch shares the same architecture as the diffusion model up to the middle block and then uses 3D convolutional projection layers to feed partial shape's $\mathbf{c}$ intermediate representation into the diffusion model in a spatially consistent manner. 
\cref{tab:parameters} shows the hyperparameters chosen in our experiments.

\begin{figure}[htpb]
    \centering
    \includegraphics[width=\linewidth]{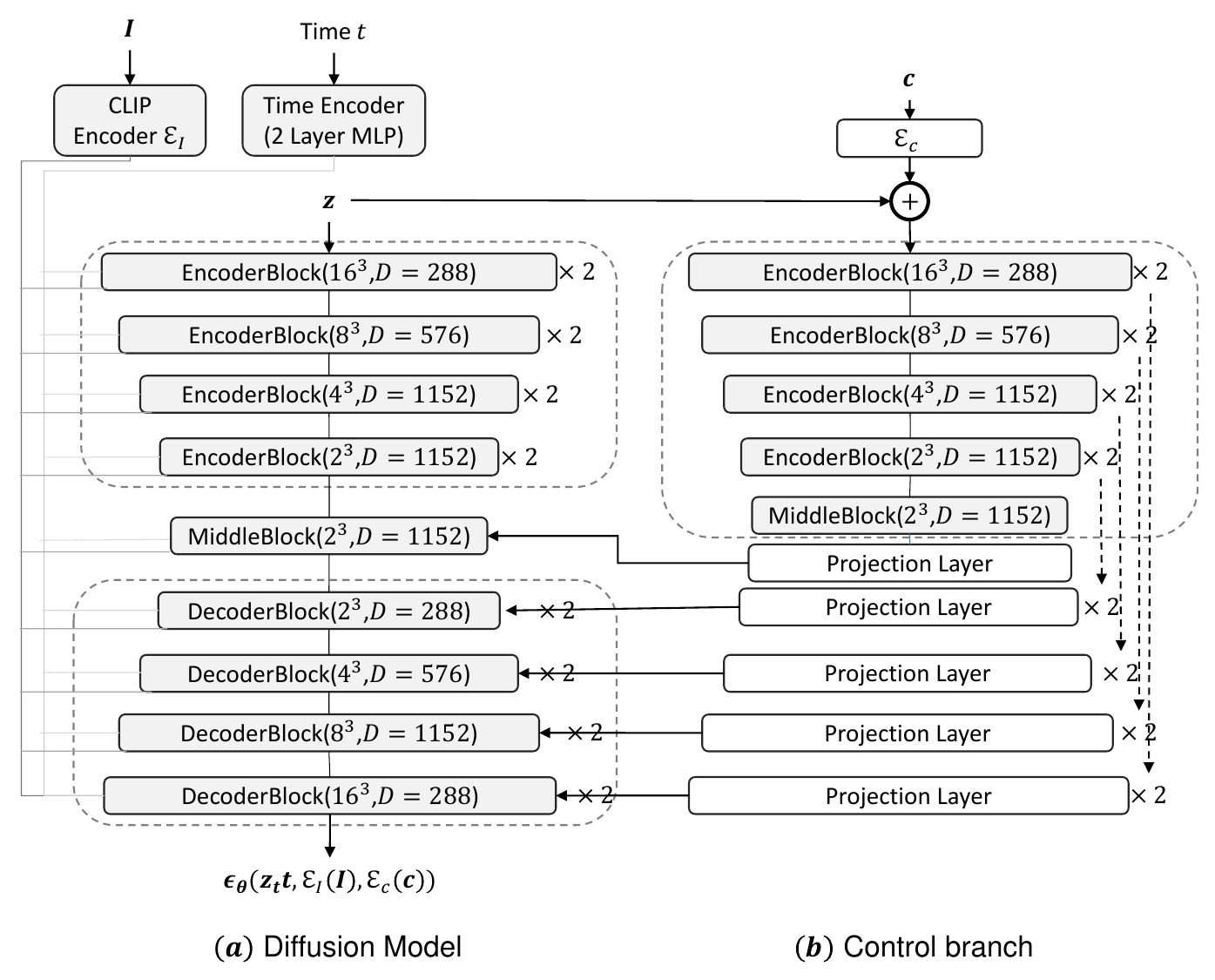}
    \caption[Network Architecture]{Our diffusion model's architecture (a) is connected with a Control branch (b). Both encoder and decoder contain eight blocks with two spatial transformers at resolutions $2^3$ and $4^3$. Each transformer contains several attention and self-attention mechanisms.}
    \label{fig:architecture}
\end{figure}

\section{More Qualitative Examples}
\begin{figure}[ht]
    \raggedleft
    \begin{overpic}[width=\linewidth, trim=-1300 0 0 0, clip]{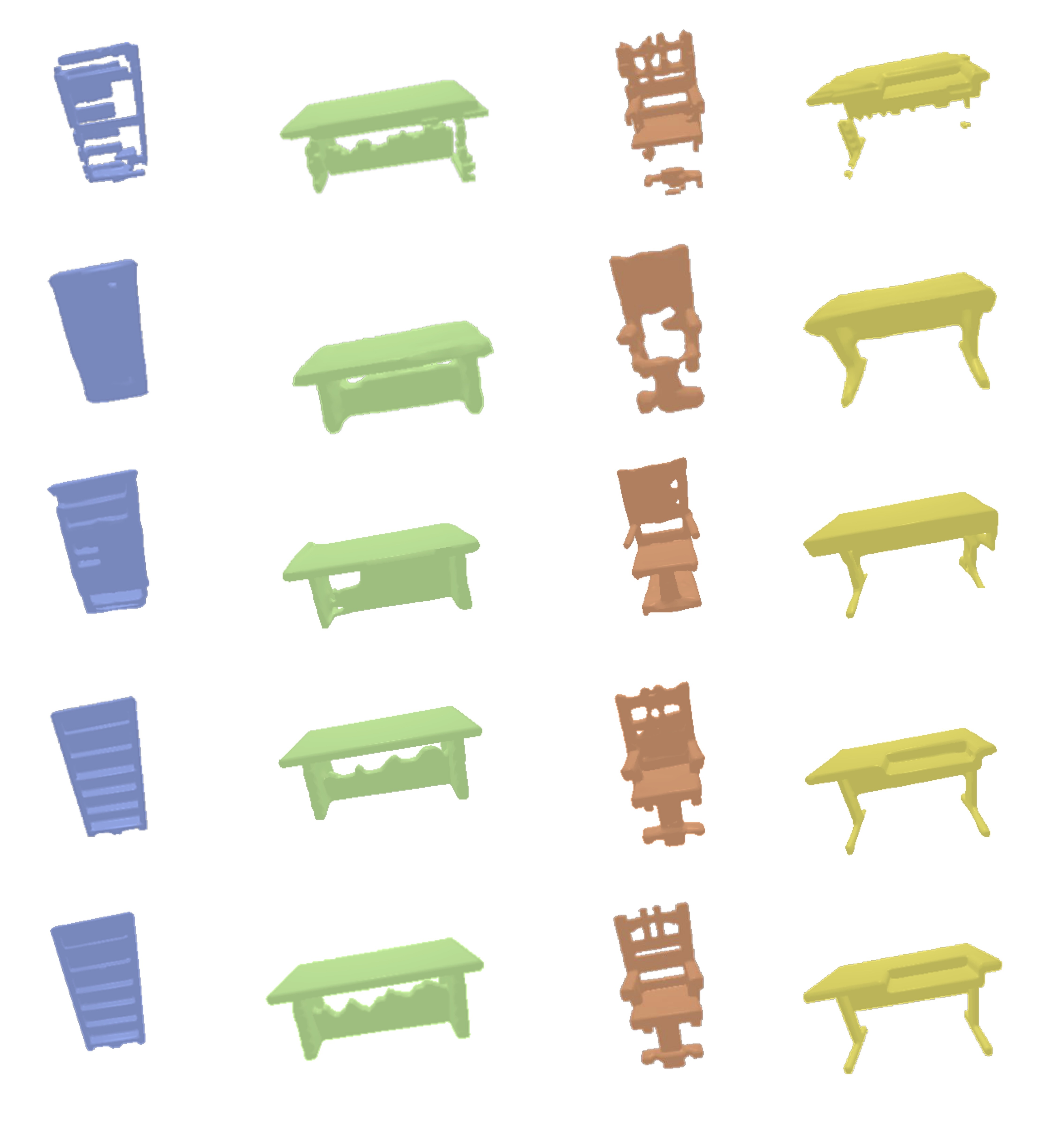}
        \put(10, 75){\makebox[0pt]{Input}}
        \put(10, 58){\makebox[0pt]{3D-EPN~\cite{dai2017_3depn}}}
        \put(10, 42){\makebox[0pt]{AutoSDF~\cite{mittal2022autosdf}}}
        \put(10, 27){\makebox[0pt]{Ours}}
        \put(10, 10){\makebox[0pt]{GT}}
    \end{overpic}
    \caption{Qualitative comparison of our method against state-of-the-art approaches on several samples from the ShapeNet dataset~\cite{chang2015shapenet}. Compared to prior methods, our approach produces plausible, non-oversmoothed shapes while recovering fine details, such as the notch in the table (rightmost column).}
    \label{fig:results_3depn2}
\end{figure}

\begin{figure}[ht]
    \centering
    \small
    \begin{overpic}[width=\linewidth, trim=0 0 0 -350, clip]{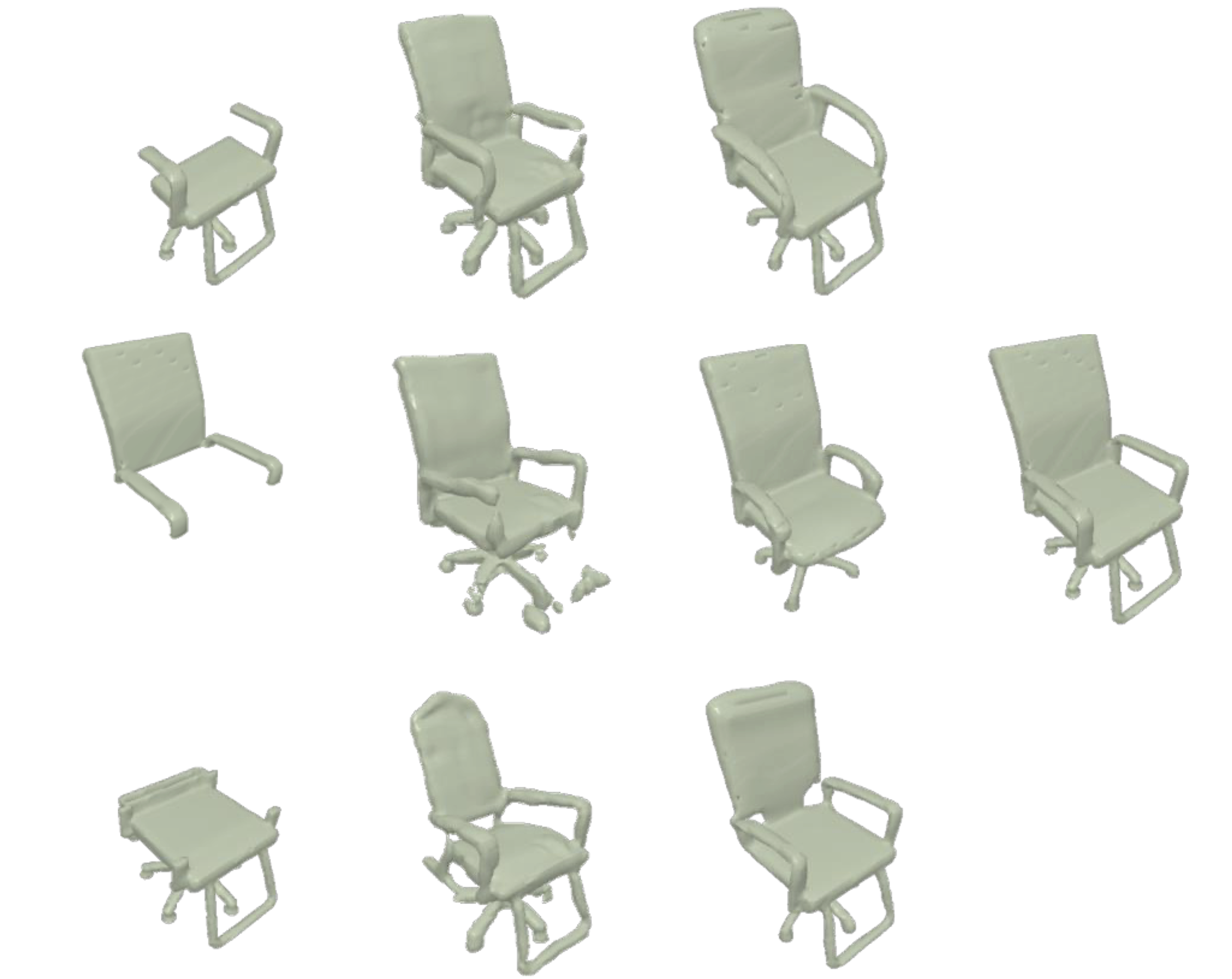}
        \put(15, 83){\makebox[0pt]{Input}}
        \put(39, 83){\makebox[0pt]{SDFusion~\cite{cheng2023sdfusion}}}
        \put(63, 83){\makebox[0pt]{Ours}}
        \put(90, 23){\makebox[0pt]{GT}}
    \end{overpic}
    \begin{overpic}[width=\linewidth, trim=0 0 0 -350, clip]{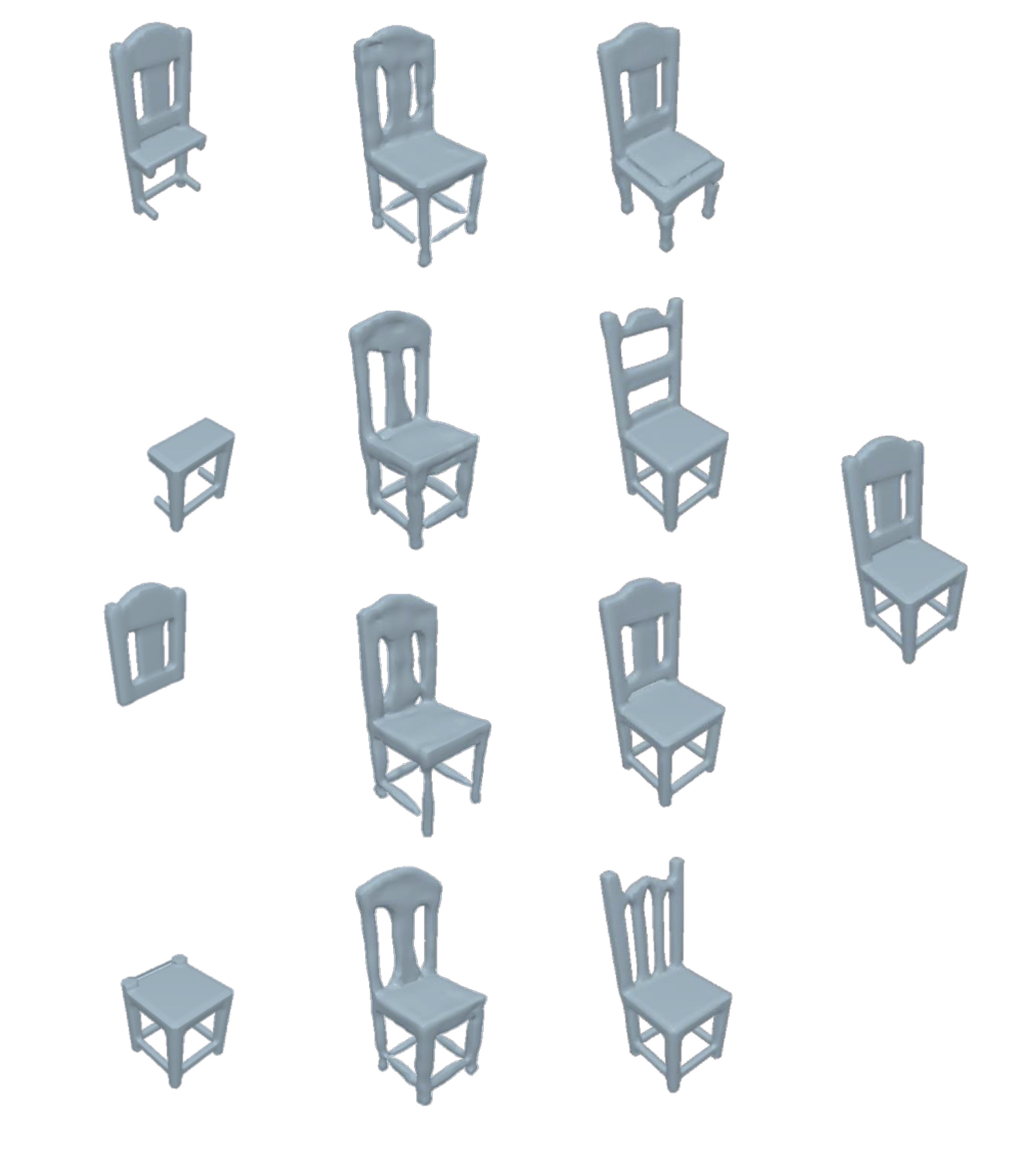}
        \put(12, 95){\makebox[0pt]{Input}}
        \put(33, 95){\makebox[0pt]{SDFusion~\cite{cheng2023sdfusion}}}
        \put(51, 95){\makebox[0pt]{Ours}}
        \put(73, 35){\makebox[0pt]{GT}}
    \end{overpic}
    \caption{Shape completion comparison with SDFusion~\cite{cheng2023sdfusion} on samples from the ShapeNet dataset~\cite{chang2015shapenet}. Our method produces more plausible and diverse results, whereas SDFusion~\cite{cheng2023sdfusion} generates noisier outputs and tends to collapse to similar shapes.}
    \label{fig:sdfusion}
\end{figure}

\begin{figure}[ht]
   \centering
   \includegraphics[width=\linewidth]{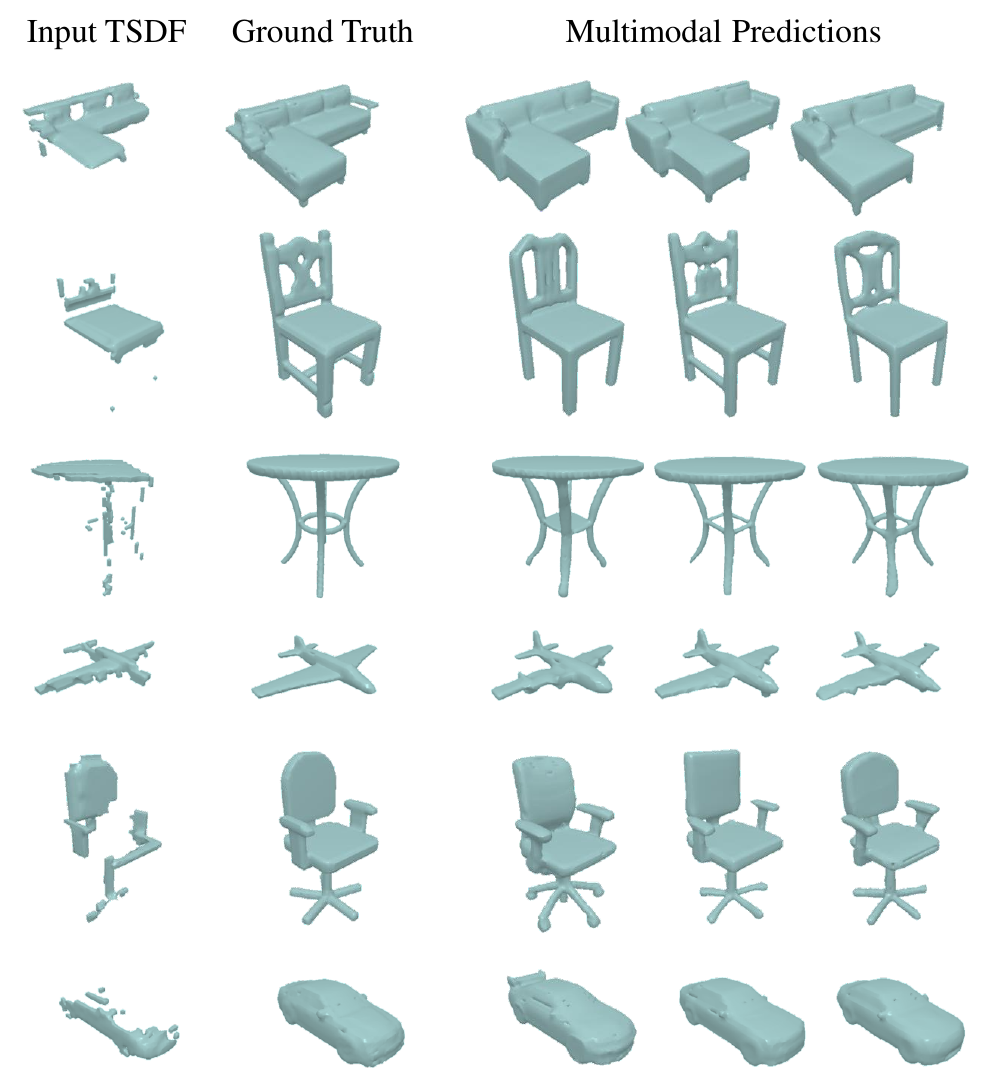}
   \caption{Additional diverse predictions generated by our model. Note that it can produce completely separate modes, such as different backrest designs in the second row.}
   \label{fig:multimodal2}
\end{figure}

\begin{figure}[ht]
   \centering
   \includegraphics[width=\linewidth]{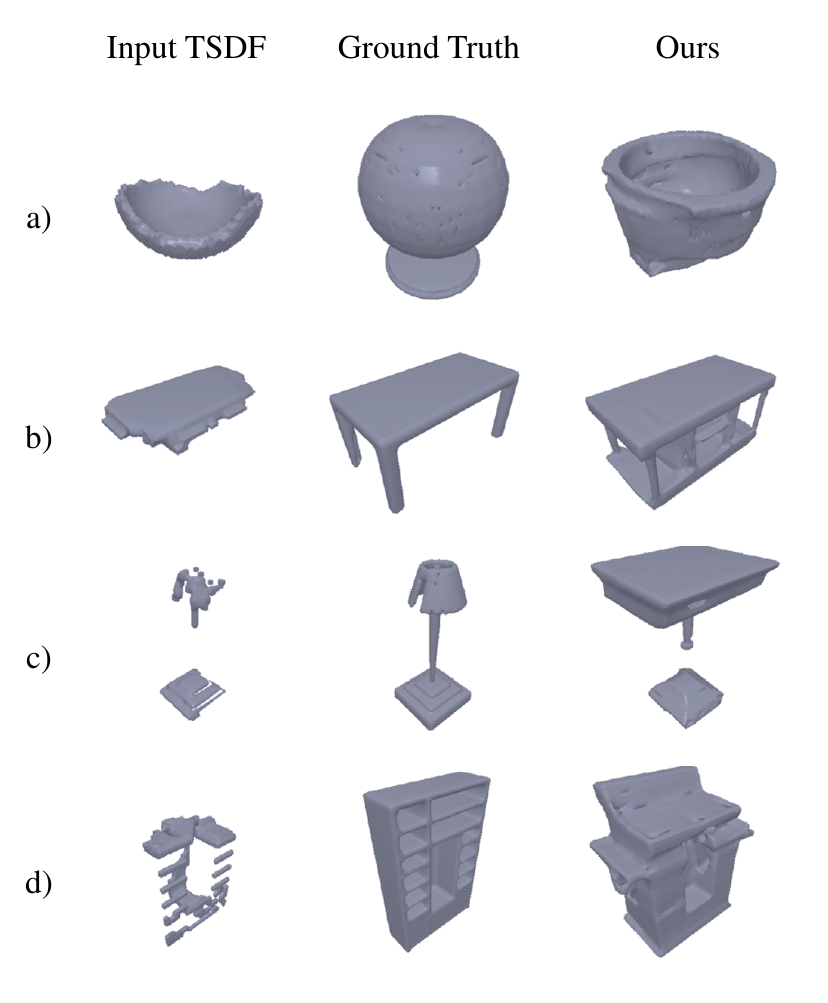}
   \caption{Examples of failure cases where our model does not accurately predict high-fidelity, realistic shapes. (a) The model predicts a container, although the ground-truth shape is a lamp. (b) The latent diffusion model unnecessarily generates a more complex shape than the ground truth. (c, d) For highly incomplete inputs, the model completes the shape as a different class—for instance, predicting a table and a sofa instead of the ground-truth lamp and cabinet.}
   \label{fig:failures}
\end{figure}

\begin{table}[htpb]
  \caption[Hyperparameters]{Values for most important hyperparameters.}\label{tab:parameters}
  \centering
  \begin{tabular}{l c c}
    \toprule
      Name & Symbol & Value \\
    \midrule
      TSDF Truncation threshold & $thresh$ & $3.0$ \\
      Resolution of the TSDF voxel grid & $S$ & $64$ \\
      Resolution of the latent space & $S_l$ & $16$ \\
      VQ-VAE latent feature size & $D$ & $3$  \\
      Codebook Size & $K_{\mathcal{Z}}$ & $16,384$ \\
      VQ-VAE 3D commitment loss weight & $\lambda$ & $0.5$ \\
      VQ-VAE 2D reconstruction loss weight & $\gamma_R$ & $0.4$ \\
      VQ-VAE 2D adversarial loss weight & $\gamma_A$ & $0.4$ \\
      Diffusion timesteps at training & $T$  & $1000$  \\
      Diffusion timesteps at inference (\cite{song2020denoising_ddim}) & $T_{inf}$  & $100$  \\
      Diffusion variance start value & $\beta_1$ & $8.5e^{-4}$ \\
      Diffusion variance end value & $\beta_T$ & $0.012$ \\
      Attention resolutions & $A_{res}$ & $[2,4]$ \\
      CLIP Features size & $D_{CLIP}$ & $768$ \\
      Batch Size & $bs$ & $8$ \\
      Learning Rate - VQ-VAE & $lr_{vqvae}$ & $1e^{-4}$ \\
      Learning Rate - Diffusion Model & $lr_{diff}$ & $2.5e^{-5}$ \\
    \bottomrule
  \end{tabular}
\end{table}